\documentclass[lettersize, journal]{IEEEtran}
\usepackage{amsmath,amsfonts}
\usepackage{algorithmic}
\usepackage{algorithm}
\usepackage{array}

\usepackage{textcomp}
\usepackage{stfloats}
\usepackage{url}
\usepackage{verbatim}
\usepackage{graphicx}

\usepackage{subcaption}

\usepackage{graphicx}
\usepackage{amsmath}
\usepackage{amssymb}
\usepackage{booktabs}
\usepackage{multirow}
\usepackage{diagbox}
\usepackage{bm}
\usepackage{bbm}
\usepackage{mathrsfs}
\usepackage{amsmath}
\usepackage{booktabs}       
\usepackage{amsfonts}       
\usepackage{nicefrac}       
\usepackage{microtype}      
\usepackage{xcolor}         
\usepackage[numbers]{natbib}

\usepackage[hidelinks]{hyperref}       
\usepackage[capitalize]{cleveref}

\begin{document}

\title{On the Privacy Effect of Data Enhancement \\ via the Lens of Memorization}

\author{Xiao~Li,~\IEEEmembership{Member,~IEEE},
        Qiongxiu~Li,~\IEEEmembership{Member,~IEEE}, 
        Zhanhao~Hu,
        and~Xiaolin~Hu$^\star$,~\IEEEmembership{Senior~Member,~IEEE}
\IEEEcompsocitemizethanks{
\IEEEcompsocthanksitem{X. Li, Q. Li, Z. Hu, and X. Hu are with the Department of Computer Science and Technology, Institute for Artificial Intelligence, BNRist, THBI, IDG/McGovern Institute for Brain Research, Tsinghua University, Beijing, 100084, China.}
\IEEEcompsocthanksitem{X. Hu is also with the Chinese Institute for Brain Research (CIBR), Beijing 100010, China. }
\IEEEcompsocthanksitem{$^\star$Corresponding author: Xiaolin Hu.} 
\IEEEcompsocthanksitem{Email: \{lixiao20, huzhanha17\}@mails.tsinghua.edu.cn; \{qiongxiuli, xlhu\}@mail.tsinghua.edu.cn.}
}
}

\maketitle

\begin{abstract}
Machine learning poses severe privacy concerns as it has been shown that the learned models can reveal sensitive information about their training data. Many works have investigated the effect of widely adopted data augmentation and adversarial training techniques, termed data enhancement in the paper, on the privacy leakage of machine learning models. Such privacy effects are often measured by membership inference attacks (MIAs), which aim to identify whether a particular example belongs to the training set or not. We propose to investigate privacy from a new perspective called \textit{memorization}. Through the lens of memorization, we find that previously deployed MIAs produce misleading results as they are less likely to identify samples with higher privacy risks as members compared to samples with low privacy risks.  To solve this problem, we deploy a recent attack that can capture individual samples' memorization degrees for evaluation. Through extensive experiments, we unveil several findings about the connections between three essential properties of machine learning models, including privacy, generalization gap, and adversarial robustness. We demonstrate that the generalization gap and privacy leakage are less correlated than those of the previous results. 
Moreover, there is not necessarily a trade-off between adversarial robustness and privacy as stronger adversarial robustness does not make the model more susceptible to privacy attacks.
\end{abstract}


\section{Introduction}
\label{sec:intro}
Due to the availability of ever-increasing datasets and computing power, we have witnessed a paradigm shift from traditional data analysis towards machine learning over the last decades. Rather than making prior assumptions, as is done in conventional modeling techniques, machine learning especially deep neural networks (DNNs) enables the system to directly learn from data.  Machine learning has achieved superior performance and has been deployed in various applications such as image processing, natural language processing, etc. As the data used for training machine learning models are often collected from local devices such as tablets and wearable devices \cite{poushter2016smartphone}, it contains sensitive personal information such as speech, images, GPS location, and medical records. It is thus crucial to protect the training data from being revealed to others. However, several studies \cite{song2017machine,carlini2019secret,zhang2021ACM,carlini2021extracting} have shown that machine learning models especially DNNs raise severe privacy concerns, as they tend to memorize sensitive information about the training data. 

To quantitatively evaluate the privacy leakage that a machine learning model reveals about its training data, a basic approach that has been intensively used is the so-called \emph{membership inference} \cite{shokri2017membership}. That is, given access to a target model, the goal of the adversary is to determine whether a particular data point was used for training this target model (being a member) or not (being a non-member). Such membership information can reveal quite sensitive information about the individuals such as the health conditions \cite{alavijeh2019quality} and serve as the basis for stronger types of privacy attacks \cite{carlini2021extracting}.

Several studies show that the attack success rates of membership inference attacks (MIAs) are highly correlated with the generalization gap, i.e., the difference between training and test accuracies~\cite{shokri2017membership, salem2018ml, song2019privacy, yeom2018privacy, leino2020stolen}.  Such correlation is also observed when applying different data enhancement methods including data augmentation and adversarial training. It has been shown in \citet{song2019privacy} that applying adversarial training can make the model more vulnerable to MIAs, and they conclude that the main reason is that the generalization gap becomes larger after applying adversarial training than standard training. The data augmentation methods, on the other hand, are widely believed to be effective in reducing the privacy leakage \cite{shokri2017membership, sablayrolles2019white, yeom2018privacy} as they are usually helpful in avoiding overfitting. Label smoothing~\cite{szegedy2016rethinking}, as a particular data augmentation method, however, is recognized to increase privacy leakage while reducing the generalization gap simultaneously~\cite{kaya2021does, hintersdorftrust}.

But the results shown in the aforementioned works might be misleading as the deployed MIAs for measuring the privacy leakage have the following limitations: 1) It has been criticized in several works \cite{rezaei2021difficulty, carlini2021membership, hintersdorftrust} that the previous MIAs often have quite high false positive rates (FPR), i.e., many non-members are falsely identified as members. However, a good attack should obtain meaningful attack rates under low FPR  regions, as it is more realistic for practical applications such as computer security \cite{kolter2006learning,lazarevic2003comparative}. As an example shown by \citet{carlini2021membership}, if an attack with overall $50.05\%$ accuracy can reliably identify just $0.1\%$ members without any false alarm, i.e., FPR=0, and judges the remaining samples by random guess with $50\%$ accuracy, it puts much more risk to the model than another attack which guesses any sample with a chance of $50.05\%$ being correct. In this case, the latter has a high FPR. 2) We find that the previous MIAs are inconsistent with the privacy risks on individual data points, even though they could have high overall success rates (see \cref{subsec.ave}). Specifically, they have more difficulties in identifying training samples with high privacy risks as members compared to the samples with low privacy risks, which is at odds with the intuition that samples with higher privacy risks should be more easily identified. 

We propose to address the above limitations by taking a new perspective called \textit{memorization}~\cite{feldman2020does,feldman2020neural}. A data point is said to be memorized if the output of the model is quite sensitive to this individual data point, e.g., the prediction confidence of the learned model on this particular data point could be quite low unless it appears in the training set \cite{feldman2020does}. The concept of memorization fundamentally captures the privacy risk under the framework of differential privacy (DP)~\cite{dwork2006, dwork2006calibrating, feldman2020does}, which is generally considered to be a strong privacy definition.  Empirically, we find that a recent attack called Likelihood Ratio Attack (LiRA) \cite{carlini2021membership} is effective in reflecting the memorization degree, as we show in \cref{subsec:CarliniMIA}. LiRA also demonstrates much better performance under the low FPR regions compared with other MIAs. Therefore, we adopt LiRA to reinvestigate the privacy effects of both data augmentation and adversarial training. 

\subsection{Paper contribution}
Through extensive investigations, we unveil several non-trivial findings (see \cref{sec:pri_gap} and \cref{sec:pri_adv} for details), which urge the community to rethink the relations among three important properties of machine learning models, including privacy leakage, generalization gap, and adversarial robustness. The major findings include:
\begin{itemize}
\item  Unlike the previous studies \cite{shokri2017membership, salem2018ml, song2019privacy, yeom2018privacy,hu2021membership} showing that the generalization gap and privacy leakage are highly correlated, our results demonstrate a much weaker correlation. 
\item Applying adversarial training can increase the memorization degrees of training samples, thereby resulting in more privacy leakage compared to models without adversarial training. However, for adversarially trained models,
stronger adversarial robustness does not necessarily come with the cost of privacy leakage.
\end{itemize}
To the best of our knowledge, this is the first systematic evaluation of data augmentation and adversarial training via the lens of memorization. 

\subsection{Outline}

The rest of the paper is organized as follows. \cref{sec:rwork} and \ref{sec:preliminary} introduce related work and necessary fundamentals, respectively.  \cref{sec:consis} explains the definition of memorization score and conducts statistical analysis to show the consistencies between the results of different MIAs and memorization scores. \cref{sec:eval} investigates the privacy effect of applying data augmentation and adversarial training on machine learning models via extensive experiments. \cref{sec:pri_gap} and \ref{sec:pri_adv} analyze the relationship between privacy and generalization gap, and between privacy and adversarial robustness, respectively. Conclusions are given in  \cref{sec:conclu}. 

\section{Related Work}
\label{sec:rwork}
This section reviews existing works on investigating the influence on privacy by applying data augmentation and adversarial training. 
\subsection{Data Augmentation and Privacy}
Due to the fact that the attack success rate of many MIAs is highly correlated with the degree of overfitting, investigating how data augmentation affects privacy leakage has received attention \cite{shokri2017membership, sablayrolles2019white,yu2021does,kaya2021does}.  As a common practice to avoid overfitting, typical data augmentation methods such as flipping and cropping are shown to be effective in mitigating MIAs  \cite{sablayrolles2019white}.  
\citet{kaya2021does} further conduct a systematic investigation by applying seven data augmentation techniques such as label smoothing, random cropping, and mixup. The results show that it is difficult to use data augmentation to achieve substantial mitigation effects against MIAs while achieving better generalization gaps.  In addition, label smoothing is shown to be able to increase both the privacy leakage and the test accuracy simultaneously \cite{kaya2021does, hintersdorftrust}.  

\subsection{Augmented Information Improves Privacy Attack} \label{subsec.mquery}
It has been shown in several studies \cite{jayaraman2021revisiting,choquette2021label,yu2021does} that exploiting the information of augmented data would help to improve the attack success rate.  MIAs can be classified into two types: augmentation-unaware and augmentation-aware attacks.  The former assumes that the adversary does not have knowledge of the augmented data but simply uses random augmentation to probe the model. It has been shown that, by querying the model multiple times, using the random augmented data generated with Gaussian noise, the attack success rate can be improved ~\cite{jayaraman2021revisiting}. The latter assumes a stronger scenario where the particular augmented data used in training is known to the adversary. \citet{choquette2021label} show that the attack success rate can be significantly improved with the knowledge of the augmented data. Moreover,  \citet{yu2021does} showed that the augmentation-aware attack can obtain a higher success rate on models trained with some data augmentation than the ones without augmentation.  

\subsection{Adversarial training  and privacy}
Adversarial training is recognized to be one of the most effective ways to improve the adversarial robustness of DNNs \cite{obf, adaptive20, rock, laat, downstream}, which is crucial in the security community. Compared to data augmentation, the relationship between adversarial training and privacy is relatively under-explored. 
It is shown in \citet{song2019privacy}  that the adversarially trained models, compared with the non-adversarially (standardly) trained models, are more susceptible to MIAs. They show that the privacy effect of adversarial training is related to several parameters including generalization gap, adversarial perturbation intensity, and model capacity. However, the MIAs used in the above-discussed studies are limited in reflecting the privacy risks of individual samples from the memorization perspective (as shown in \cref{subsec.ave}). In addition, they all do not report results by the metric under low FPR regions.

\section{Preliminaries}
\label{sec:preliminary}
In this section, we introduce the necessary fundamentals for understanding the rest of the paper.

\subsection{Deep neural network}
We consider feed-forward DNNs for classification tasks under the usual supervised setting. Suppose we have a training set $D_\mathrm{tr} = \{(x,y)|(x,y) \in \mathcal{X}\times\mathcal{Y}\}$, where $x$ is the feature vector (e.g., image) and $y$ is the corresponding label. We denote the DNN model with parameter $\theta$ as $f_{\theta}:\mathcal{X} \to \mathcal{Y}$. During training, the data augmentation $T:\mathcal{X}\times\mathcal{Y}\to\mathcal{P}$ is applied to the training data to improve its diversity, where $\mathcal{P}$ is the set of all the probability measures defined on the power set $2^{\mathcal{X}\times\mathcal{Y}}$. Together, the optimal parameter $\theta^*$ of the model is fitted by:
\begin{equation}
\begin{aligned}
\label{eq:deeplearning}
\theta^* = \mathop{\arg\min}_{\theta}{\sum_{(x, y)\in D_\mathrm{tr}}\mathbb{E}_{(\tilde{x}, \tilde{y})\sim T(x, y)}[L(f_{\theta}(\tilde{x}), \tilde{y})}],
\end{aligned}
\end{equation} 
where $L(\cdot, \cdot)$ is the loss function.

\subsection{Data Enhancement}
Since both data augmentation and adversarial training involve the process of adding certain examples into the training set to enhance the performances, they are known as data enhancement techniques. In this paper, we investigate eight popular data augmentation methods and four adversarial training methods, as described below. 

Data augmentation methods: 
\begin{enumerate}
    \item Random Cropping and Flipping: sample new features by randomly cropping and horizontally flipping patches from the original feature in the training set.
    \item Label smoothing ~\cite{szegedy2016rethinking}: replace the hard labels with the soft continuous labels by uniformly assigning probabilities to other classes. Therefore, the probability of the augmented label is $\tilde{p_i} = 1-\frac{(n-1)\epsilon}{n}$ for $i=y$ and it is $\tilde{p_i} = \frac{\epsilon}{n}$ for $i\neq y$, where $\epsilon\in(0, 1)$ and $n$ denotes the number of the classes.
    \item Disturblabel ~\cite{xie2016disturblabel}: change a portion of ground-truth (GT) labels to incorrect labels, namely, $\tilde{y}=\epsilon y+(1-\epsilon) y_{f}$, where $\epsilon$ is randomly sampled from $\{0, 1\}$ and $y_f \in \{1,2,...,n\}\setminus \{y\}$ denotes the incorrect label.
    \item Gaussian Noise ~\cite{cohen2019certified}: add Gaussian noise to each feature. The new feature $\tilde{x} = x + \epsilon$, where $\epsilon\sim \mathcal{N}(0, \sigma^2 I)$.
    \item Cutout ~\cite{devries2017improved}: mask out a random square area of size $M \times M$ from each feature.
    \item Mixup ~\cite{zhang2018mixup}: blend two features $x_0, x_1$ by a random ratio $\gamma$ and creates a new feature $\tilde{x} = \gamma x_0 + (1-\gamma) x_1$. The corresponding label is $\tilde{y} = \gamma y_0 + (1-\gamma) y_1$. 
    \item Jitter ~\cite{krizhevsky2012imagenet}: randomly change the brightness, contrast, saturation, and hue of each image.
    \item Distillation ~\cite{hinton2015distilling}: train an auxiliary DNN $\hat{f}$ with the original training set and use the auxiliary DNN's soft outputs and temperature $T$ as GT labels of the training features when training the target DNN. The temperature $T$ determines the flatness of the soft labels.
\end{enumerate}

Adversarial training techniques: 
\begin{enumerate}
    \item PGD-AT \cite{madry2018towards}: use PGD attack to generate adversarial examples $x_{\mathrm{adv}}$ based on original features and replaces the original feature with the adversarial examples at each iteration of the training, i.e., $\tilde{x}=x_{\mathrm{adv}}$.
    \item TRADES \cite{zhang2019theoretically}: use PGD attack to generate adversarial examples $x_{\mathrm{adv}}$, too. It differs from PGD-AT in that its loss function consists of two components: $L(f_{\theta}(x), y) + L(f_{\theta}(x), f_{\theta}(x_{\mathrm{adv}}))/\lambda$. The first component is the same as the loss of the standard training while the second component encourages the model to treat $x$ and $x_{\mathrm{adv}}$ equally. The two components are weighted by $\lambda$. In the framework of \cref{eq:deeplearning}, the discrete distribution of the transformation $T(x, y)$ can be represented as $\mathrm{Pr}(x, y)=\frac{\lambda}{\lambda + 1}$ and $\mathrm{Pr}(x, f_{\theta}(x_{\mathrm{adv}}))=\frac{1}{\lambda+1}$, where $\mathrm{Pr}(\cdot)$ indicates the probability.
    \item AWP \cite{awp}: use regularization to explicitly flatten the weight loss landscape of PGD-AT by a double-perturbation mechanism.
    \item TRADES-AWP \cite{awp}: incorporate the regularization mechanism of AWP into the TRADES method.
\end{enumerate}

\subsection{Membership Inference Attack}
The goal of MIA is to identify whether a specific data sample was used in training a particular model or not.  MIA has become one of the most widely investigated privacy attacks due to its simplicity. Many existing MIA approaches \cite{shokri2017membership,salem2018ml,yeom2018privacy,long2018understanding,jia2019memguard} can achieve high attack accuracy by exploiting the fact that machine learning models often behave differently to the data used or not used for training. For example, the model is often more confident about the training data than the test data. Thus, by setting a threshold to certain features such as the loss, confidence score, entropy, etc, the attack can achieve high accuracy in distinguishing members from non-members. Such kinds of approaches are referred to as metric-based approaches as they simply deploy a preset threshold to decide the membership.  Another type of MIA approach is the classifier-based approach, which involves training a binary classifier to distinguish members and non-members.  An effective technique to train such classifiers is known as shadow training \cite{shokri2017membership}.  The main idea of shadow training is to train several so-called shadow models, which are trained similarly to the target model to mimic its behavior.  The classifier-based approaches tend to be more computationally complex compared to the metric-based methods.

\begin{figure*}[!t]
  \centering
  \begin{subfigure}[b]{0.245\linewidth}
    \includegraphics[width=\linewidth]{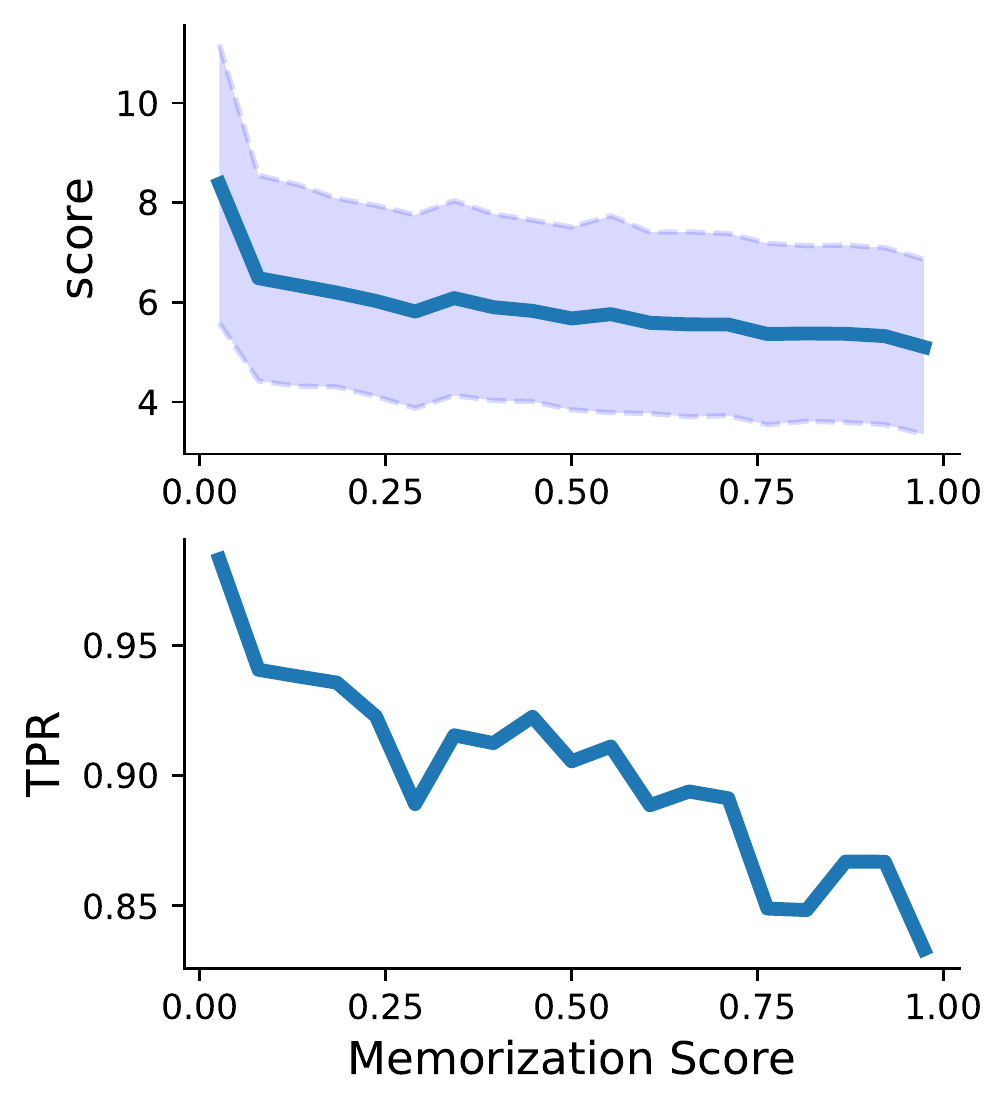}
    \caption{MaxPreCA}
    \label{fig.mem_conf}
  \end{subfigure}
  \begin{subfigure}[b]{0.245\linewidth}
    \includegraphics[width=\linewidth]{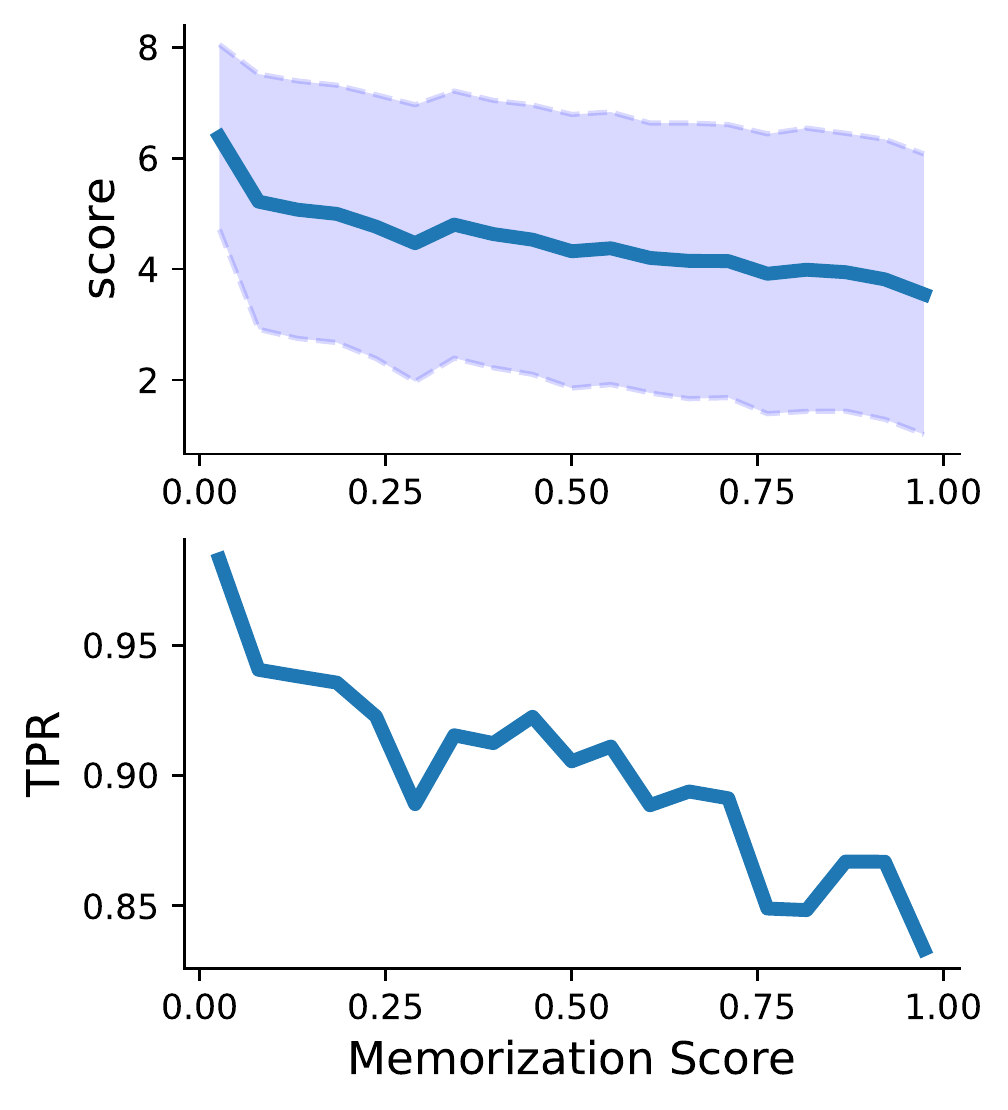}
    \caption{Loss}
    \label{fig.mem_loss}
  \end{subfigure}
  \begin{subfigure}[b]{0.245\linewidth}
    \includegraphics[width=\linewidth]{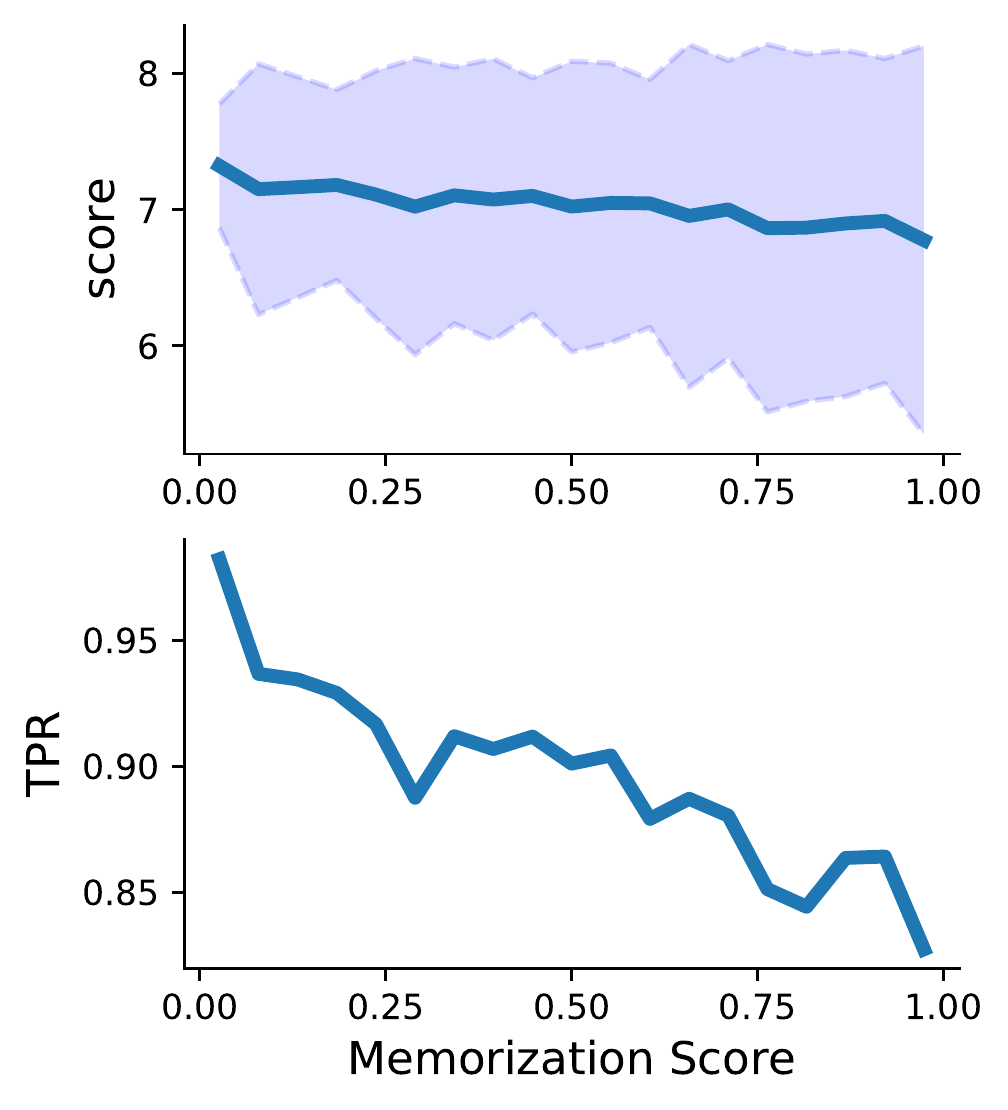}
    \caption{Modified entropy}
    \label{fig.mem_mh}
  \end{subfigure}
  \begin{subfigure}[b]{0.245\linewidth}
    \includegraphics[width=\linewidth]{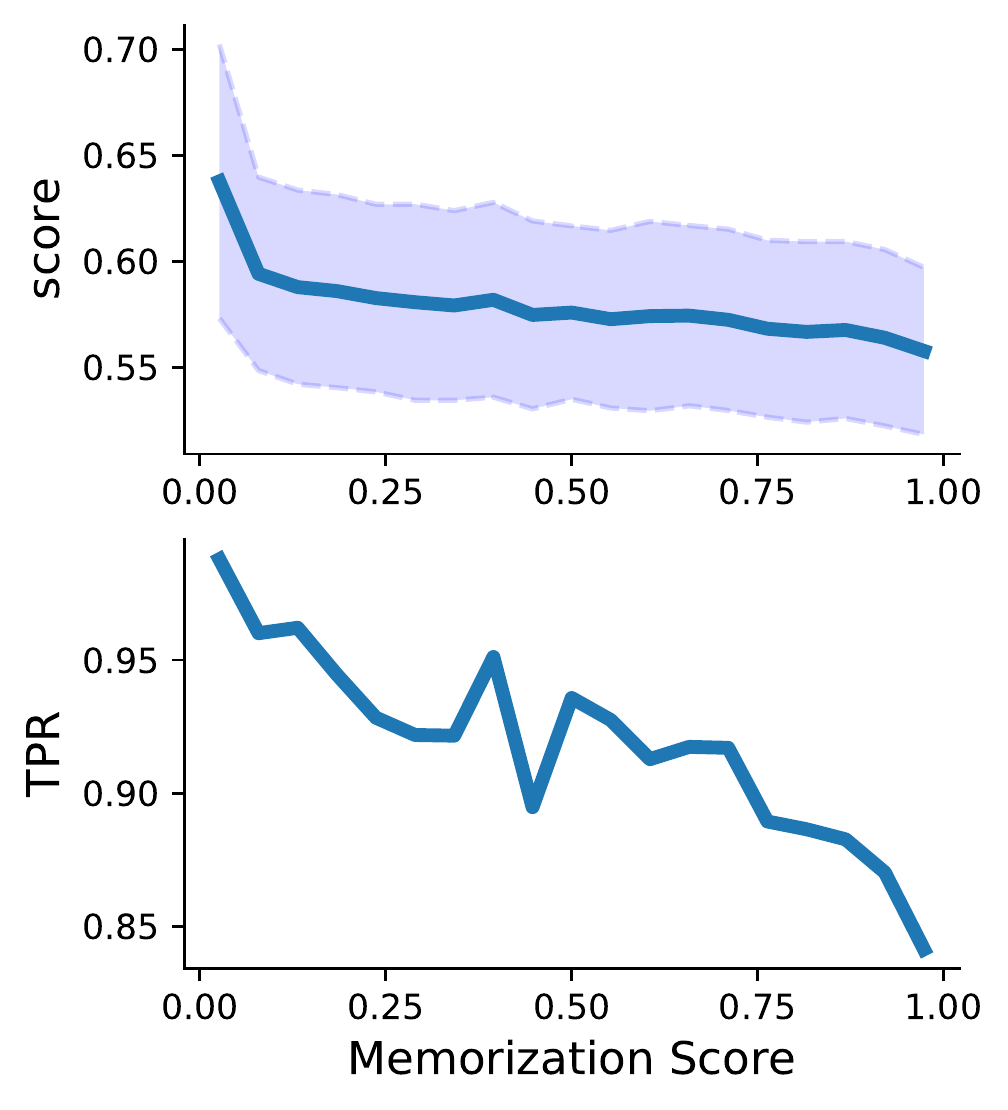}
    \caption{Binary classifier}
    \label{fig.mem_shadow}
  \end{subfigure}
  \begin{subfigure}[b]{0.245\linewidth}
    \includegraphics[width=\linewidth]{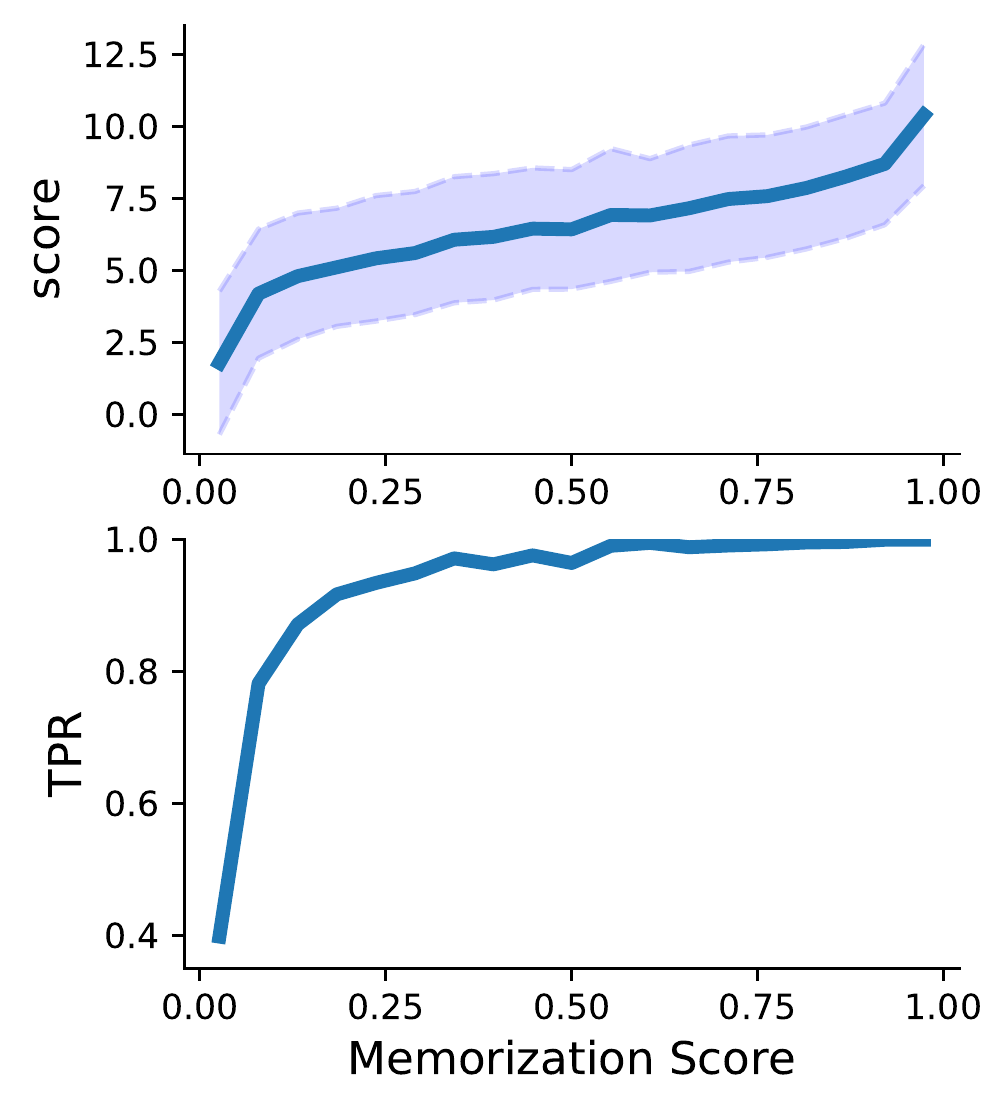}
    \caption{Bayes calibrated loss}
    \label{fig.mem_tau}
  \end{subfigure}
  \begin{subfigure}[b]{0.245\linewidth}
    \includegraphics[width=\linewidth]{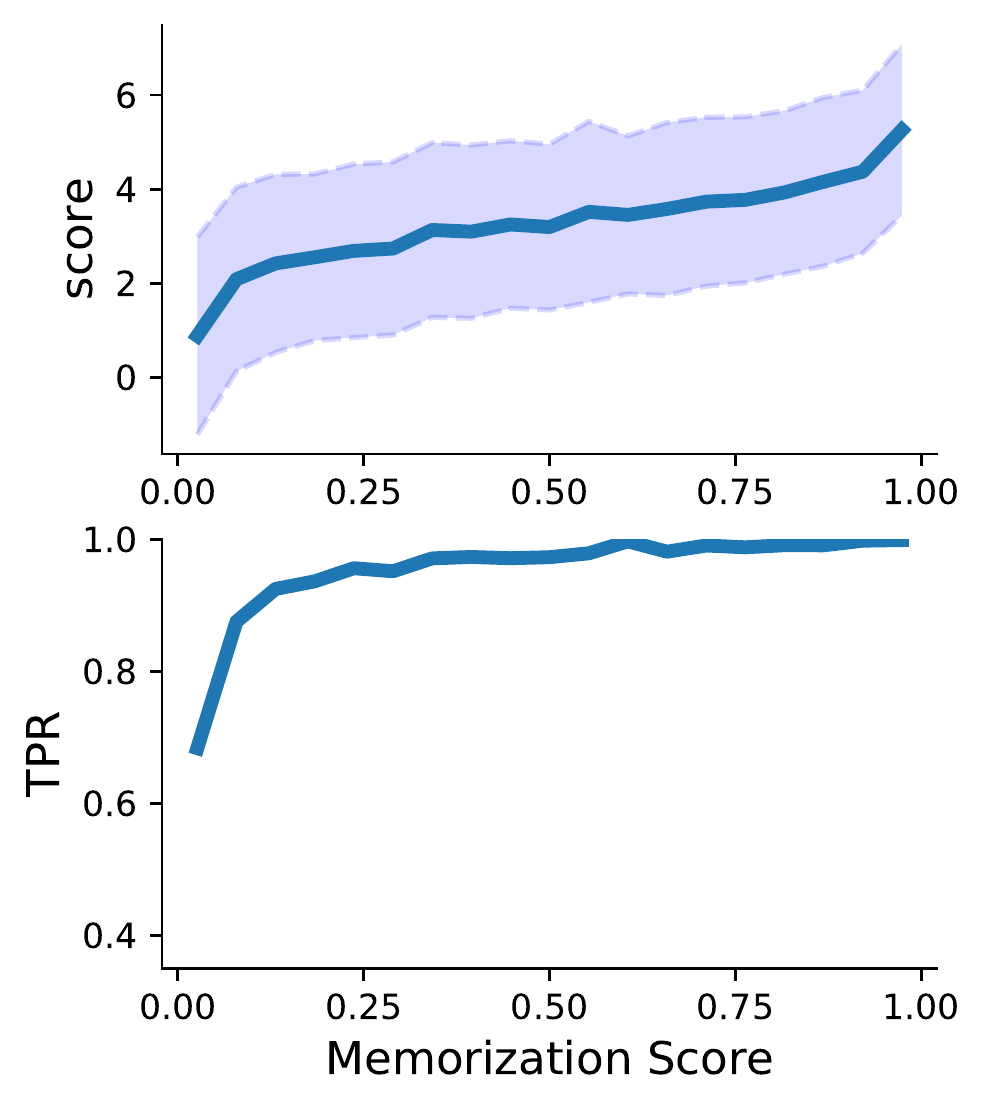}
    \caption{Difficulty calibrated loss}
    \label{fig.mem_halftau}
  \end{subfigure}
    \begin{subfigure}[b]{0.245\linewidth}
    \includegraphics[width=\linewidth]{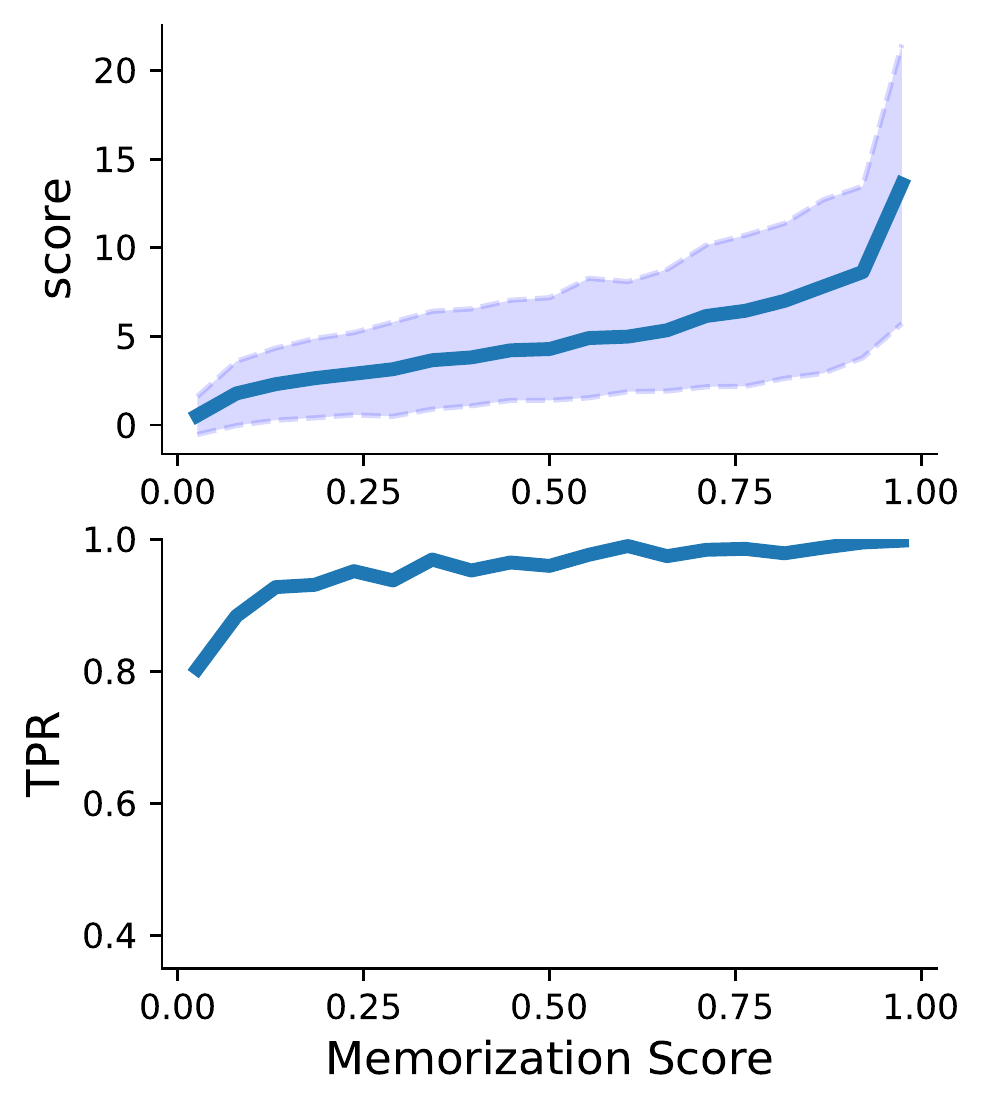}
    \caption{LiRA}
    \label{fig.mem_base}
  \end{subfigure}
  \caption{Feature score (top) and TPR (bottom) of each bin in terms of memorization score for a target model trained on CIFAR-100 using (a) MaxPreCA \cite{salem2018ml}, (b) Loss attack \cite{yeom2018privacy}, (c) Modified entropy attack \cite{song2021systematic}, (d) Binary classifier \cite{shokri2017membership}, (e) Bayes calibrated loss \cite{sablayrolles2019white}, (f) Difficulty calibrated loss \cite{importance}, and (g) LiRA \cite{carlini2021membership}.}
  \label{fig.mem}
\end{figure*}

%

\section{Consistency of MIAs and Memorization}
\label{sec:consis}
In this section, our objective is to explain why existing research might yield misleading outcomes owing to two primary limitations and to demonstrate how adopting a memorization perspective can address these issues. To achieve this, we start by introducing the concept and definition of memorization scores. We then investigate the consistency between results obtained from different MIAs and the memorization scores. 
At last, we emphasize the need to reevaluate the privacy risks associated with data enhancements, using the lens of memorization scores to gain a more accurate understanding.

\subsection{Memorization}
A data point is said to be memorized by the model if it has a high impact on the model's behavior. To ensure that such impact is solely caused by this particular sample, one often needs to use the \textit{leave-one-out} setting. Namely, except for this particular sample all other settings are the same. As an example, \citet{feldman2020does} defines a memorization score which measures how much information about the label of an individual data sample is being memorized by the model. Specifically,  given the training set $D_\mathrm{tr}$ and the learning algorithm  $\mathcal{A}$, for an arbitrary  sample $(x,y)\in D_\mathrm{tr}$, its (label) memorization score is defined as:
\begin{equation}
\begin{aligned}
\label{eq.labelm}
&\operatorname{mem}(\mathcal{A}, D_\mathrm{tr},(x,y))  \\
&=\underset{f_{\theta} \sim \mathcal{A}(D_\mathrm{tr})}{\operatorname{Pr}}\left[f_{\theta}(x)=y\right]-\underset{f_{\theta} \sim \mathcal{A}(D_\mathrm{tr} \setminus (x,y))}{\operatorname{Pr}}\left[f_{\theta}(x)=y\right],
\end{aligned}
\end{equation}
where $D_\mathrm{tr} \setminus (x,y)$ denotes the dataset $D_\mathrm{tr}$ with the sample $(x,y)$ being removed.  
This definition is shown effective as it can assign atypical examples or outliers with high memorization scores and typical or easy samples with low memorization scores on various datasets including the CIFAR-100 \cite{cifar} and ImageNet \cite{krizhevsky2012imagenet} datasets \cite{feldman2020neural}.  This complies perfectly with the intuition that an atypical example or outlier is often at a higher privacy risk, as the model will behave quite differently on it when it is in or out of the training set.

We now proceed to investigate whether the results produced by existing MIAs are consistent with memorization scores.  Given that the memorization scores of the CIFAR-100 dataset have been pre-computed by \citet{feldman2020neural}, where $4000$ models were trained in total, we deploy it as the ground truth memorization scores for the following evaluations. Without loss of generality, we select representative approaches including both metric-based and classifier-based MIAs. Here we trained 128 ResNet-18 \cite{resnet} models with half of the data points randomly selected from 60,000 data points from CIFAR-100 (see the Hyper-Parameters paragraph in \cref{sec:setting} for detailed training configurations). For each model, the other half was used as the non-member set for conducting MIA. Then we randomly chose ten trained models as the target model. The remaining 127 models were recognized as shadow models for each respective target model (if the MIA used shadow models). For each MIA, we used the optimum threshold that maximized the overall balanced accuracy, i.e., the proportion of correctly identified samples (both members and non-members) out of the total number of samples in the dataset\footnote{Unless otherwise specified, all MIA thresholds mentioned in this work when evaluating balanced accuracy follow this setting.}. The final attack results were averaged over the ten target models. Below we present the results of consistency check, evaluating whether the attack results are consistent with memorization scores.

\subsection{Example MIAs Obtaining Low Consistency with Memorization Scores} 
\label{subsec.ave} 
Many MIAs exploit the fact that models are overconfident about the training data \cite{hintersdorf2022to}.  Such confidences may be captured by different features, e.g., the prediction confidence \cite{shokri2017membership,salem2018ml}, loss \cite{yeom2018privacy}, and so-called modified entropy \cite{song2021systematic}. That is, compared to the test data, the model usually has a higher confidence score, smaller loss, and lower entropy (uncertainty)  on the training data. For example, the main idea of the widely-adopted  maximum predication confidence-based attack \cite{salem2018ml} (called MaxPreCA in this paper) is to classify samples with the maximum prediction confidence higher than a given threshold as members, otherwise as non-members:
\begin{equation}
\begin{aligned}
    \mathcal{M}_{\text{conf}}(x, y)=\mathbbm{1}\{\max_{c}~[f_{\theta}(x)_{c}] \geq \tau\}
\end{aligned}
\end{equation}
where $\mathbbm{1}$ denotes the indicator function, $f_{\theta}(x)_{c}$ denotes the model's confidence to be category $c$, and $\mathcal{M}_{\text {conf }}(x, y)$  denotes the adversary's decision on the membership. Similarly, the decisions of popular loss attack \cite{yeom2018privacy} is given by:
\begin{equation}
\begin{aligned}
    \mathcal{M}_{\text{loss}}(x, y)=\mathbbm{1}\{-L(f_{\theta}(x), y) \geq \tau\},
\end{aligned}
\end{equation}
and the  modified entropy attack \cite{song2021systematic} is given by:
\begin{equation}
\begin{aligned}
    \mathcal{M}_{\text {m-entr}}(x, y)=\mathbbm{1}\{ [1 - f_{\theta}(x)_{y}]\log[f_{\theta}(x)_{y}] \\
    + \sum_{c\neq y}f_{\theta}(x)_{c}\log[1-f_{\theta}(x)_{c}] \geq \tau\},
\end{aligned}
\end{equation}
note that $f_{\theta}(x)_{y}$ denotes the model's confidence of $x$ on the correct label $y$. Here we also investigate the classifier-based MIA, using the shadow training method proposed in \citet{shokri2017membership}. More specifically, for each target model, we used ten shadow models and the logits of these models and sample labels as features to train a multi-layer perceptron (MLP) classifier for binary membership classification. We denote this as a ``binary classifier'' approach.

To investigate if these attacks can capture the privacy risk of individual data points, in \cref{fig.mem}(a-d) we demonstrate their feature scores, i.e., the maximum prediction confidence, the loss, the modified entropy and the prediction score of binary-classifier (top panel), and the True Positive Rate (TPR) (bottom panel) versus the memorization score \cite{feldman2020neural}. We first divide all the samples into $20$ bins according to their memorization score. The top panel shows the averaged feature scores (the solid blue line) of the samples in different bins along with their standard deviation (purple shadow). For visualization, the feature scores of the top panels in \cref{fig.mem}(b-c) are transformed using exponential scaling (monotonic function), ensuring that they are on a similar scale as the other panels. The bottom panel shows the TPR of each bin for each MIA. It is obvious that the higher the memorization score is, the less probable it gets identified as members correctly. Therefore, the results of these attacks have a low consistency with memorization scores. 

\subsection{Example MIAs Obtaining High Consistency with Memorization Scores}
\label{subsec:CarliniMIA} 
Based on the definition in \cref{eq.labelm}, MIAs that exhibit a high degree of consistency with memorization scores should effectively capture how a model's behavior varies when a specific sample is included or excluded from the training set. In this context, we highlight three example MIAs that leverage such behavioral differences: the Bayes calibrated loss approach \cite{sablayrolles2019white}, the difficulty calibrated loss approach \cite{importance}, and LiRA \cite{carlini2021membership}. Given that LiRA is essentially an evolved form of the earlier two methods, our discussion will begin with an in-depth explanation of LiRA, and then we will provide a brief description of other approaches.

LiRA considers the distribution of the model's prediction on an individual data point when it is in or out of the training set. It requires training a number of shadow models such that for each sample $(x,y)$, half models include it in the training set and the other half models do not, denoted as IN and OUT models, respectively. Denote the sets of scaled confidences of sample $(x,y)$ computed using IN and OUT models as $\mathcal{Q}_{\mathrm{in}}$ and $\mathcal{Q}_{\mathrm{out}}$,  respectively:
\begin{equation}
\label{eq.inout}
\begin{aligned}
\mathcal{Q}_{\mathrm{in}}=\{\phi(f_{\theta}(x)_{y}): (x,y)\in D_{\mathrm{tr}}\} \\
\mathcal{Q}_{\mathrm{out}}=\{\phi(f_{\theta}(x)_{y}): (x,y)\notin D_{\mathrm{tr}}\},
\end{aligned}
\end{equation}
where $\phi(p)=\log \left(\frac{p}{1-p}\right)$. $ \mathcal{Q}_{\mathrm{in}}$ and $\mathcal{Q}_{\mathrm{out}}$ are used to fit two Gaussian distributions, denoted as IN distribution $\mathcal{N}\left(\mu_{\mathrm{in}}, \sigma_{\mathrm{in}}^{2}\right)$ and OUT distribution $\mathcal{N}\left(\mu_{\mathrm{out}}, \sigma_{\text {out }}^{2}\right)$, respectively. Given an arbitrary sample, a standard likelihood-ratio test is performed to determine which distribution it more likely belongs to, where the likelihood ratio  $\Lambda$ is defined as:
\begin{equation} 
\begin{aligned} \label{eq.lamda}
\Lambda=\frac{\operatorname{Pr}\left(\phi(f_{\theta}(x)_{y}) \mid \mathcal{N}\left(\mu_{\mathrm{in}}, \sigma_{\text {in }}^{2}\right)\right)}{\operatorname{Pr}\left(\phi(f_{\theta}(x)_{y}) \mid \mathcal{N}\left(\mu_{\mathrm{out}}, \sigma_{\text {out }}^{2}\right)\right)}.
\end{aligned}
\end{equation}
A sample will be classified as a member if $\Lambda$ is higher than a threshold when evaluating balanced accuracy. 

The Bayes calibrated loss method \cite{sablayrolles2019white} is similar to LiRA except that it does not fit the output of shadow models using Gaussian. Instead, it directly uses the average output from the IN and OUT models for calibrating the sample loss. More precisely, the calibrated loss attack is given by\footnote{Note that since we assume the output is confidence, thus in this work we use the confidence score instead of loss.}:
\begin{equation}
\begin{aligned}
\mathcal{M}_{\text {bc}}(x, y) = \mathbbm{1}\{f_{\theta}(x)_{y} - \mu(x, y) \geq \tau\}
\end{aligned}
\end{equation}
where $\mu(x,y) = (\mu_{\mathrm{in}}(x,y) + \mu_{\mathrm{out}}(x,y))/2$ and $\mu_{\mathrm{in}}$ and $\mu_{\mathrm{out}}$ are the means computed like those in LiRA.

The difficulty calibrated loss approach \cite{importance} can be seen as an offline version of the above Bayes calibrated loss approach, which uses the difficulty of samples, measured by OUT models, to calibrate the loss, i.e., $\mu(x,y) = \mu_{\mathrm{out}}(x,y)$. In \cref{fig.mem}(e-g), we present the Bayes calibrated loss, the difficulty calibrated loss, and the likelihood ratio \( \Lambda \) in \cref{eq.lamda} (top panel), along with the TPR (bottom panel), against the memorization scores. The results clearly show that samples with higher memorization scores are more likely to be accurately identified as members. 

\begin{table}[!t]
  \centering
  \small
  \setlength{\tabcolsep}{2.5pt}
  {
    \begin{tabular}{c|c|c}
Method & Balanced Acc & TPR @ 0.1\% FPR \\
\hline
MaxPreCA \cite{salem2018ml} & $75.62$ & $0.08$ \\
Loss \cite{yeom2018privacy} & $75.48$ & $0.14$ \\
Modified entropy \cite{song2021systematic} & $75.66$ & $0.21$ \\
Binary classifier \cite{shokri2017membership} & $76.50$ & $0.32$ \\
Bayes calibrated loss \cite{sablayrolles2019white} & $79.75$ & $4.95$ \\
Difficulty calibrated loss \cite{importance} & $76.42$ & $19.09$ \\
LiRA \cite{carlini2021membership} & $\bm{80.31}$ & $\bm{25.71}$ \\
\end{tabular}
    
    }
   \caption{The attack success rate of different MIAs on CIFAR-100.}
  \label{tab:tpracc}
\end{table}

\subsection{Consistency Results and Low FPR Metrics}
As mentioned in the introduction and several previous studies \cite{rezaei2021difficulty, carlini2021membership, hintersdorftrust}, a prevalent limitation of many MIAs is their tendency to exhibit high FPRs. This issue becomes apparent when examining the consistency of these attack results with memorization scores. By inspecting \cref{fig.mem}(a-d), we identify two key observations. First, these attacks only achieve exceptionally high TPRs (e.g., above 0.95) within a narrow range of memorization scores. This implies that aiming for an overall low FPR will reduce the overall TPR across all regions significantly. Second, the highest TPRs are typically found in regions with very low memorization scores. This implies that aiming for an overall low FPR tends to exclude those members that present a higher privacy risk, which is also counter-intuitive.
This phenomenon can be related to the fact that samples with high memorization scores are often more complex or atypical, making them more challenging for the model to fit \cite{feldman2020neural}. Consequently, these samples tend to exhibit lower confidence compared to simpler, typical (test) samples. Therefore, implementing a stricter threshold to reduce the FPR, i.e., to prevent false identification of non-member test samples, will inadvertently exclude training samples with high memorization scores.

In contrast, MIAs with high consistency in memorization scores, as shown in \cref{fig.mem}(e-g), maintain very high TPRs across most regions, especially for those regions with high memorization scores.  As a result, these MIAs can sustain high overall TPRs even in seeking low FPR scenarios, with the identified members being those with high memorization scores, which is a logical and expected outcome. To support these findings, \cref{tab:tpracc} presents the performance of these MIAs, evaluating both the balanced accuracy and TPR at a low FPR. It is evident that MIAs with low consistency in memorization scores exhibit poor performance in low FPR regions, while those with high consistency show markedly better results. 

Among the MIAs with high consistency with memorization scores, as shown in \cref{fig.mem}(e-g),
we note that LiRA outperforms the other methods, especially in accurately classifying samples with low memorization scores. This superiority is attributed to LiRA's utilization of second-order statistics through Gaussian fitting, rather than solely relying on mean information like the other methods. Consequently, LiRA not only achieves the highest consistency with memorization scores but also the most effective attack success rate.  

\begin{table*}[!t]
  \centering
  \small
\renewcommand{\arraystretch}{1.15}{
  \setlength{\tabcolsep}{4pt}
  {
    \begin{tabular}{c|cc|cccc}
    \textbf{Method} & \textbf{Training Acc}  & \textbf{Test Acc} & \textbf{TPR @ 0.1\% FPR} & \textbf{TPR @ 0.001\% FPR} & \textbf{Log-scale AUC} & \textbf{MIA Balanced Acc} \\
    
     \hline
Base & $100.0 \pm 0.0$ & $92.8 \pm 0.2$     & $8.20 \pm 0.45$ & $2.45 \pm 0.93$     & $0.815 \pm 0.007$     & $63.34 \pm 0.26$ \\
\hline
Smooth & $100.0 \pm 0.0$ & $92.9 \pm 0.3$     & $5.22 \pm 0.66$ & $0.14 \pm 0.07$     & $0.734 \pm 0.012$     & $62.28 \pm 0.86$ \\
Disturblabel & $99.9 \pm 0.0$ & $92.7 \pm 0.3$     & $5.88 \pm 0.83$ & $0.70 \pm 0.45$     & $0.775 \pm 0.013$     & $61.69 \pm 0.24$ \\
Noise & $100.0 \pm 0.0$ & $92.6 \pm 0.2$     & $\bm{8.33} \pm 0.26$ & $\bm{2.79} \pm 0.68$     & $\bm{0.819} \pm 0.004$     & $\bm{63.56} \pm 0.24$ \\
Cutout & $100.0 \pm 0.0$ & $93.1 \pm 0.4$     & $7.71 \pm 0.39$ & $\bm{2.48} \pm 1.03$     & $0.811 \pm 0.010$     & $63.23 \pm 0.26$ \\
Mixup & $99.7 \pm 0.1$ & $93.0 \pm 0.2$     & $5.17 \pm 0.51$ & $1.31 \pm 0.40$     & $0.779 \pm 0.008$     & $60.05 \pm 0.53$ \\
Jitter & $100.0 \pm 0.0$ & $92.7 \pm 0.2$     & $\bm{8.24} \pm 0.35$ & $\bm{2.97} \pm 0.76$     & $\bm{0.819} \pm 0.004$     & $\bm{63.41} \pm 0.31$ \\
Distillation & $99.9 \pm 0.0$ & $93.2 \pm 0.2$     & $7.04 \pm 0.33$ & $2.19 \pm 0.70$     & $0.805 \pm 0.005$     & $61.57 \pm 0.39$ \\
\hline
PGD-AT & $99.2 \pm 0.1$ & $82.2 \pm 0.2$     & $\bm{23.78} \pm 0.89$ & $\bm{10.52} \pm 2.30$     & $\bm{0.897} \pm 0.005$     & $\bm{78.82} \pm 0.37$ \\
TRADES & $96.2 \pm 0.2$ & $80.0 \pm 0.4$     & $\bm{17.88} \pm 1.56$ & $\bm{8.14} \pm 1.12$     & $\bm{0.881} \pm 0.006$     & $\bm{77.21} \pm 0.65$ \\

AWP & $93.2 \pm 2.0$ & $82.6 \pm 0.9$     & $\bm{10.58} \pm 3.48$ & $\bm{3.06} \pm 1.81$     & $\bm{0.828} \pm 0.045$     & $\bm{72.13} \pm 3.76$ \\
TRADES-AWP & $91.9 \pm 0.5$ & $80.5 \pm 0.2$     & $\bm{12.43} \pm 0.89$ & $\bm{3.48} \pm 1.36$     & $\bm{0.848} \pm 0.006$     & $\bm{74.86} \pm 0.80$ \\
    \end{tabular}
    
    }
    }
  \caption{Attack success rates of different data enhancement on CIFAR-10. The 2nd and 3rd columns show the training and test accuracies of each method, respectively. The 4th - 7th columns show four metrics to evaluate the extent of privacy leakage. We highlight the MIA success rates for different data augmentation and adversarial training methods that are larger than that for Base. }
  \label{tab:c10}
\end{table*}

\subsection{Discussion on Previous Evaluations of Data Enhancements}
The findings discussed above raise significant concerns because many studies examining the privacy risks associated with data enhancements have relied on MIAs that show low consistency with memorization scores. For example, \citet{salem2018ml,yeom2018privacy,kaya2021does,yu2021does} utilize either confidence or loss-based MIAs to assess the privacy impact of data augmentation. Similarly,  \citet{song2019privacy} employs the confidence-based MIA to evaluate the risks associated with adversarial training. We suspect that the conclusions of these studies might be misleading. Therefore, it is important to reinvestigate the privacy risks associated with data augmentation and adversarial training, employing MIAs that have high consistency with memorization scores for a more accurate evaluation.

Given the superior performance of  LiRA in obtaining the highest consistency with memorization scores among the MIAs we examined, we will deploy LiRA as the primary tool to investigate how various data augmentation and adversarial training techniques impact privacy in subsequent discussions.

\begin{table*}[!t]
  \centering
  \small
\renewcommand{\arraystretch}{1.15}{
  \setlength{\tabcolsep}{4pt}
  {
    \begin{tabular}{c|cc|cccc}
    \textbf{Method} & \textbf{Training Acc}  & \textbf{Test Acc} & \textbf{TPR @ 0.1\% FPR} & \textbf{TPR @ 0.001\% FPR} & \textbf{Log-scale AUC} & \textbf{MIA Balanced Acc} \\
    
     \hline
Base & $100.0 \pm 0.0$ & $70.3 \pm 0.3$     & $34.17 \pm 1.05$ & $17.24 \pm 2.93$     & $0.922 \pm 0.002$     & $83.17 \pm 0.24$ \\
\hline
Smooth & $100.0 \pm 0.0$ & $72.2 \pm 0.4$     & $\bm{39.21} \pm 1.25$ & $\bm{19.88} \pm 3.86$     & $\bm{0.932} \pm 0.004$     & $\bm{86.35} \pm 0.22$ \\
Disturblabel & $98.0 \pm 0.2$ & $69.9 \pm 0.3$     & $19.53 \pm 0.64$ & $6.54 \pm 2.58$     & $0.879 \pm 0.007$     & $76.65 \pm 0.28$ \\
Noise & $100.0 \pm 0.0$ & $69.7 \pm 0.3$     & $33.83 \pm 0.91$ & $\bm{18.31} \pm 2.78$     & $\bm{0.923} \pm 0.003$     & $\bm{83.26} \pm 0.13$ \\
Cutout & $100.0 \pm 0.0$ & $70.3 \pm 0.3$     & $\bm{34.71} \pm 1.58$ & $\bm{17.25} \pm 5.02$     & $\bm{0.923} \pm 0.005$     & $\bm{83.53} \pm 0.22$ \\
Mixup & $99.7 \pm 0.1$ & $71.2 \pm 0.4$     & $32.73 \pm 1.13$ & $\bm{19.18} \pm 2.48$     & $0.922 \pm 0.003$     & $82.39 \pm 0.50$ \\
Jitter & $100.0 \pm 0.0$ & $70.3 \pm 0.3$     & $\bm{34.19} \pm 0.90$ & $\bm{18.37} \pm 3.62$     & $\bm{0.924} \pm 0.003$     & $\bm{83.35} \pm 0.17$ \\
Distillation & $99.8 \pm 0.0$ & $72.6 \pm 0.3$     & $28.70 \pm 0.83$ & $14.58 \pm 2.29$     & $0.911 \pm 0.002$     & $79.46 \pm 0.14$ \\
	\hline
PGD-AT & $99.5 \pm 0.0$ & $51.3 \pm 0.3$     & $\bm{68.63} \pm 0.88$ & $\bm{47.85} \pm 4.26$     & $\bm{0.972} \pm 0.001$     & $\bm{93.62} \pm 0.10$ \\
TRADES & $98.0 \pm 0.3$ & $49.0 \pm 0.5$     & $\bm{60.23} \pm 0.87$ & $\bm{37.45} \pm 5.15$     & $\bm{0.963} \pm 0.002$     & $\bm{92.19} \pm 0.23$ \\
AWP & $85.3 \pm 0.8$ & $54.4 \pm 0.3$     & $\bm{39.92} \pm 2.40$ & $\bm{17.34} \pm 4.22$     & $\bm{0.931} \pm 0.006$     & $\bm{88.10} \pm 0.38$ \\
TRADES-AWP & $95.9 \pm 0.6$ & $51.3 \pm 0.4$     & $\bm{57.51} \pm 2.02$ & $\bm{35.57} \pm 3.73$     & $\bm{0.960} \pm 0.002$     & $\bm{91.92} \pm 0.36$ \\
    \end{tabular}
    }
    }
  \caption{Attack success rates of different data enhancement on CIFAR-100. The same conventions are used as in \cref{tab:c10}. }
  \label{tab:c100}
\end{table*}

\section{Evaluating Privacy Effects of Data Enhancement}\label{sec:eval}
We now proceed to evaluate the privacy effects of both data augmentation and adversarial training. Through the experiments, we aim to investigate the relations of privacy, adversarial robustness, and generalization gap, as they are all crucial properties for a machine learning model. 
\subsection{Experimental Settings}
\label{sec:setting}
We used 32 NVIDIA 3080 GPUs to perform the experiments. The code for the experiments is implemented by \texttt{Pytorch} and is available at \url{https://github.com/LixiaoTHU/privacy_and_aug}.

\vspace{2mm}
\noindent\textbf{Dataset.}
Following previous work \cite{carlini2021membership, advmia, kaya2021does}, we used the CIFAR-10 \cite{cifar} and CIFAR-100 \cite{cifar} datasets for MIA evaluations. We additionally used the SVHN \citep{svhn} dataset for a comprehensive evaluation. Both CIFAR-10 and CIFAR-100 contain 60,000 natural images with a resolution of 32 $\times$ 32 from 10 and 100 categories, respectively. SVHN contains 73,257 color images of numbers for training, with the resolution of $32 \times 32$, and the first 60,000 training images from SVHN are used in this work.

\vspace{2mm}
\noindent\textbf{MIA Settings.}
In this work, we assume a black-box setting where the adversary only has query access to the outputs of the target model on given samples. Without loss of generality, here we assume the output is the prediction confidence and the shadow models are trained using the same data enhancement method as the target model.
We used LiRA to measure the privacy leakage of different data augmentation and adversarial training methods. We trained 128 models for each data enhancement method. Each model of the 128 models used roughly 30,000 data points as the training samples (members) and the remaining roughly 30,000 data points as the test samples (non-members). For each data point, we guaranteed that 64 out of the 128 models were the IN models, and the remaining 64 models were the OUT models.
All models used the same training recipes except for the data enhancement strategies. 
For evaluation, we randomly selected 10 out of the 128 models as target models. The remaining 127 models were recognized as shadow models for each respective target model. After that, we performed the MIA evaluation on target models with all 60,000 data points. The MIA performance results were reported as the mean and standard deviation for each attack metric across the ten target models.

\vspace{2mm}
\noindent\textbf{Models.}
Unless otherwise specified, we used the standard ResNet-18 \cite{resnet} for the experiments on CIFAR-10 and CIFAR-100 and a vanilla convolutional neural network (CNN) for the experiments on SVHN. The CNN we used contained six convolutional layers with output channel numbers (256, 512, 512, 512, 512, 512), kernel sizes (3, 2, 3, 3, 2, 3), strides (1, 2, 1, 1, 2, 1), and padding sizes (1, 0, 1, 1, 0, 0). Batch normalization \cite{bn} was applied to each layer. After the image of $32 \times 32$ was down-sampled by these layers to be $6 \times 6$, an average pooling operation was followed. Two fully connected layers, which can be seen as part of a MLP network, were added at the end of the output, with sizes 200 and 10, respectively. We denoted it as CNN-8 in the following text. Due to computational resource constraints, we did not conduct the same experiments on larger architectures.

\vspace{2mm}
\noindent\textbf{Hyper-Parameters.}
Each model was optimized by stochastic gradient descent with an initial learning rate of 0.1 and a momentum of 0.9 for 100 epochs on a single GPU. Multi-step decay which scales the learning rate by 0.1 was used on the 75th and 90th epochs. The batch size is set to 256.

The hyper-parameters of each data augmentation method were set to achieve relatively high test accuracy by searching (see Appendix \ref{supp:a} for details). Unless other specified, for all adversarial training methods, we set the maximal perturbation $\epsilon$ under $\ell_{\infty}$ norm to be 8. We set the step size to be $\epsilon / 8$ and the number of iterative steps to be 10. In addition, following the default setting of each method, the regularization parameter $\lambda$ was set to be $1/6$ for TRADES, the perturbation intensity $\gamma$ was set to be $1 \times 10^{-2}$ for AWP, and $5 \times 10^{-3}$ for TRADES-AWP.

\begin{figure*}[!t]
  \centering
  \includegraphics[width=0.9\linewidth]{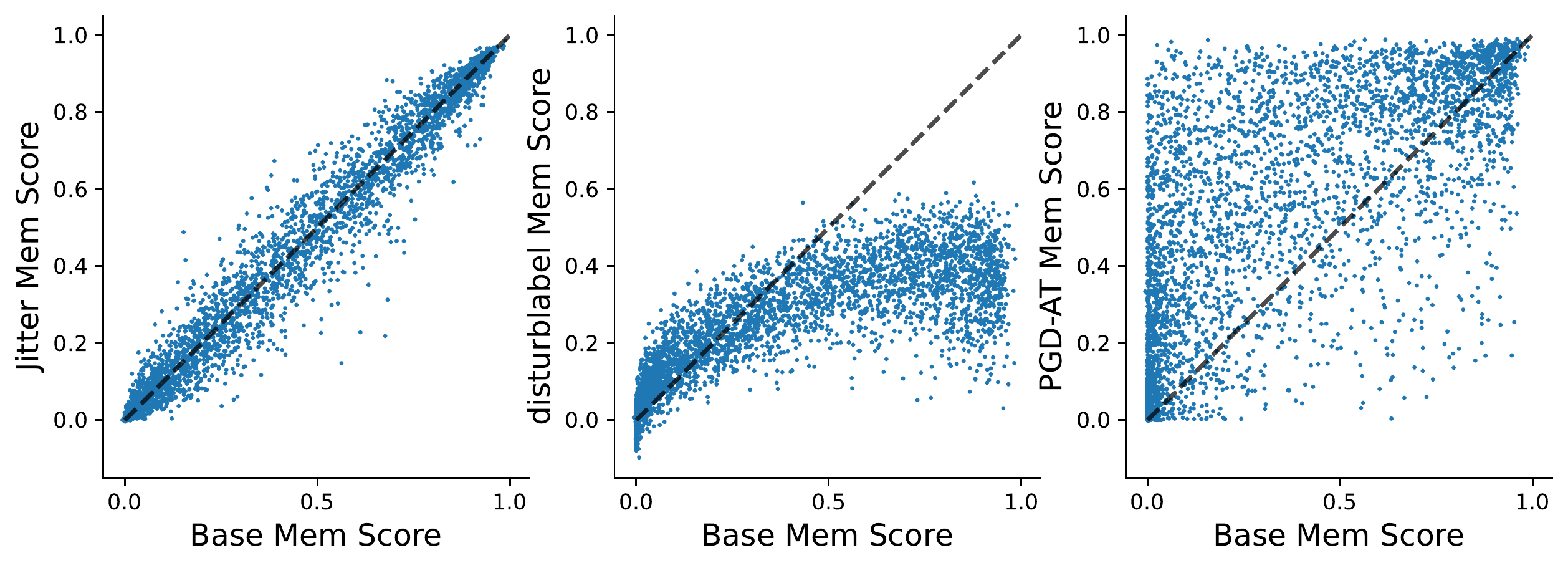}
  \caption{Memorization scores of 5,000 randomly selected samples using Jitter (top), Disturblabel (middle) and PGD-AT model (bottom) v.s., Base model. }
\label{fig.memCons}
\end{figure*}

\subsection{Evaluation Results}
Tables \ref{tab:c10} and \ref{tab:c100} respectively show the training and test accuracies and the MIA results by multiple queries on CIFAR-10 and CIFAR-100. The results on SVHN are shown in Appendix \ref{supp:b}. We denote Random Cropping and Flipping by the \textit{Base} method. Different from most previous studies \cite{kaya2021does, yu2021does, song2019privacy}, we evaluated the privacy leakage of seven data augmentation methods from label smoothing to Distillation and four adversarial training methods (\cref{sec:preliminary}) combined with Base. This is a practical setting as Random Cropping and Flipping has now become a default setting in the computer vision field and it often brings considerable improvements in test accuracy. It has been criticized in \citet{rezaei2021difficulty} that the models with low test accuracies are not practically useful for evaluating privacy leakage. See the results of different data augmentation methods without Base in Appendix \ref{supp:b} as examples, where the test accuracy of Base exceeds those of other data augmentation methods by at least $8.4\%$ on CIFAR-10 and $12.6\%$ on CIFAR-100. Unless otherwise specified, all data augmentation and adversarial training methods also use Base as default.

With LiRA, we evaluated all models using four metrics: TPR @ $0.1\%$ FPR, TPR @ $0.001\%$ FPR, Log-scale Area Under the Curve (AUC), and the Balanced Accuracy. The numbers after $\pm$ denote the standard deviations. 
As mentioned earlier, it is more reasonable to use the metric of TPR under low FPR regions for evaluating privacy leakage. However, on the one hand, we empirically found that the results of TPR @ $0.001\%$ FPR were unstable since their standard deviations were relatively large. On the other hand, $0.001\%$ FPR might be too strict as it has been shown in \citet{feldman2020neural} that both CIFAR-10 and CIFAR-100 contain quite a few pairs of hard samples that are very similar. These samples will inevitably be misclassified with high confidence as members when their counterparts are in the member set, thereby causing some false positive cases and resulting in FPRs that exceed the $0.001\%$ tolerance. Therefore, we mainly use TPR @ $0.1\%$ FPR as the attack success rate for the following analysis.

\subsection{Evaluation Results and Memorization Scores}
To verify whether our evaluation results indeed reflect the degree of memorization, in \cref{fig.memCons} we compare the memorization scores of 5,000 randomly selected samples computed using the same method as in \citet{feldman2020neural} for three cases on CIFAR-100: Jitter, Disturblabel, and PGD-AT v.s. Base. Wherein the attack success rates are similar to, lower than, and higher than Base, respectively. Clearly, the corresponding changes in the memorization scores are consistent with the attack success rates. The memorization scores of the samples for Base and Jitter are similar (lying around the diagonal), which explains the similar attack success rate against the two methods. The memorization scores of many samples for Disturblabel are lower than for Base, especially the samples with high memorization scores for Base. Therefore, one reason why Disturblabel reduces privacy leakage is that it can reduce the memorization scores of many atypical samples. The memorization scores for PGD-AT are in general higher than those for Base (the points are distributed on the upper of the diagonal). Thus, one major reason why adversarial training causes a higher privacy leakage is that it memorizes many training samples that are not memorized by standardly trained models. Overall, we conclude that our evaluation results are reliable as they can reflect the degree of memorization.

\begin{figure*}[!t]
\centering
\begin{subfigure}[b]{0.3\linewidth}
  \centering
  \includegraphics[height=200pt]{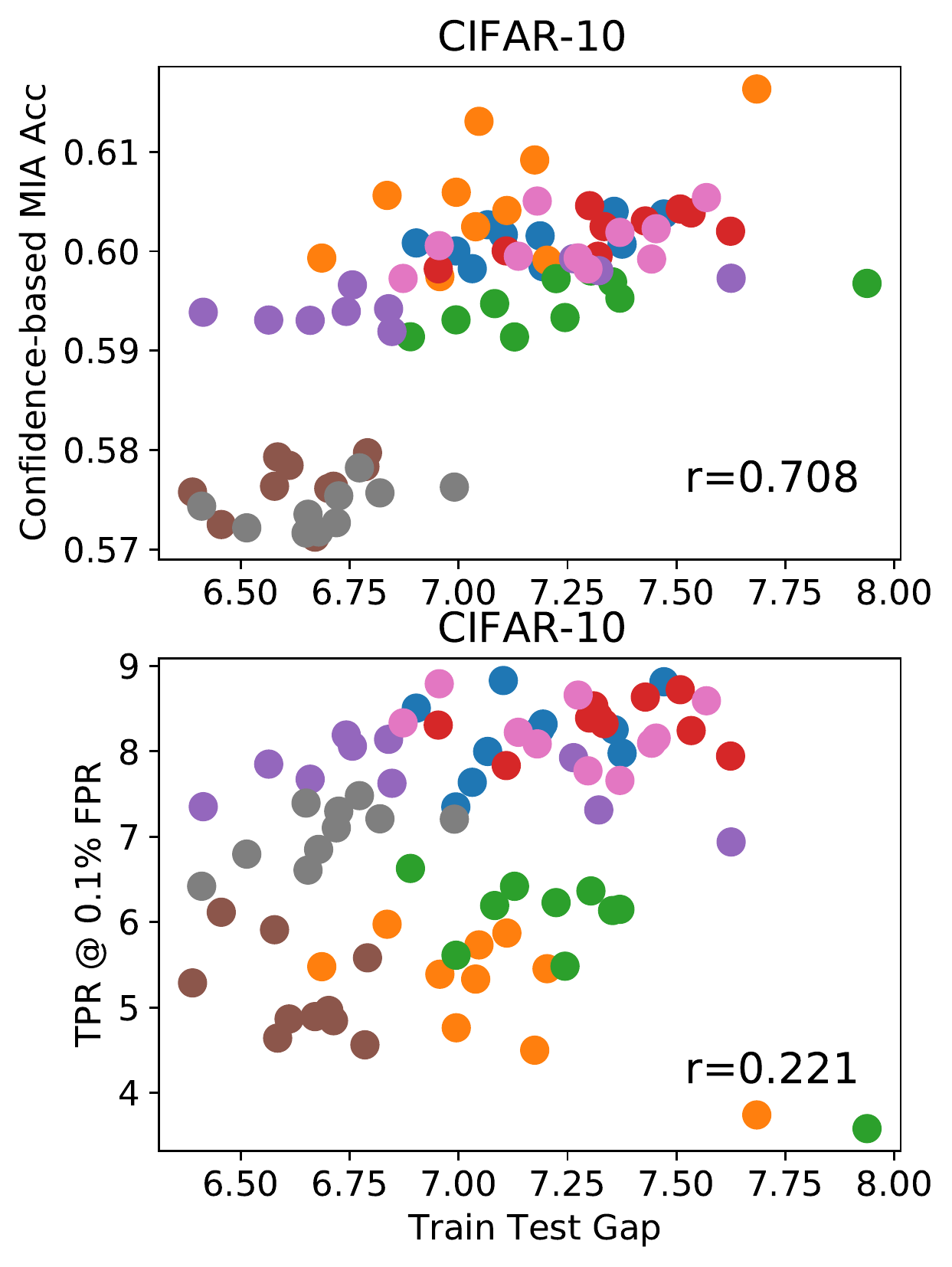}
  \caption{CIFAR-10}
\end{subfigure}
\begin{subfigure}[b]{0.3\linewidth}
  \includegraphics[height=200pt]{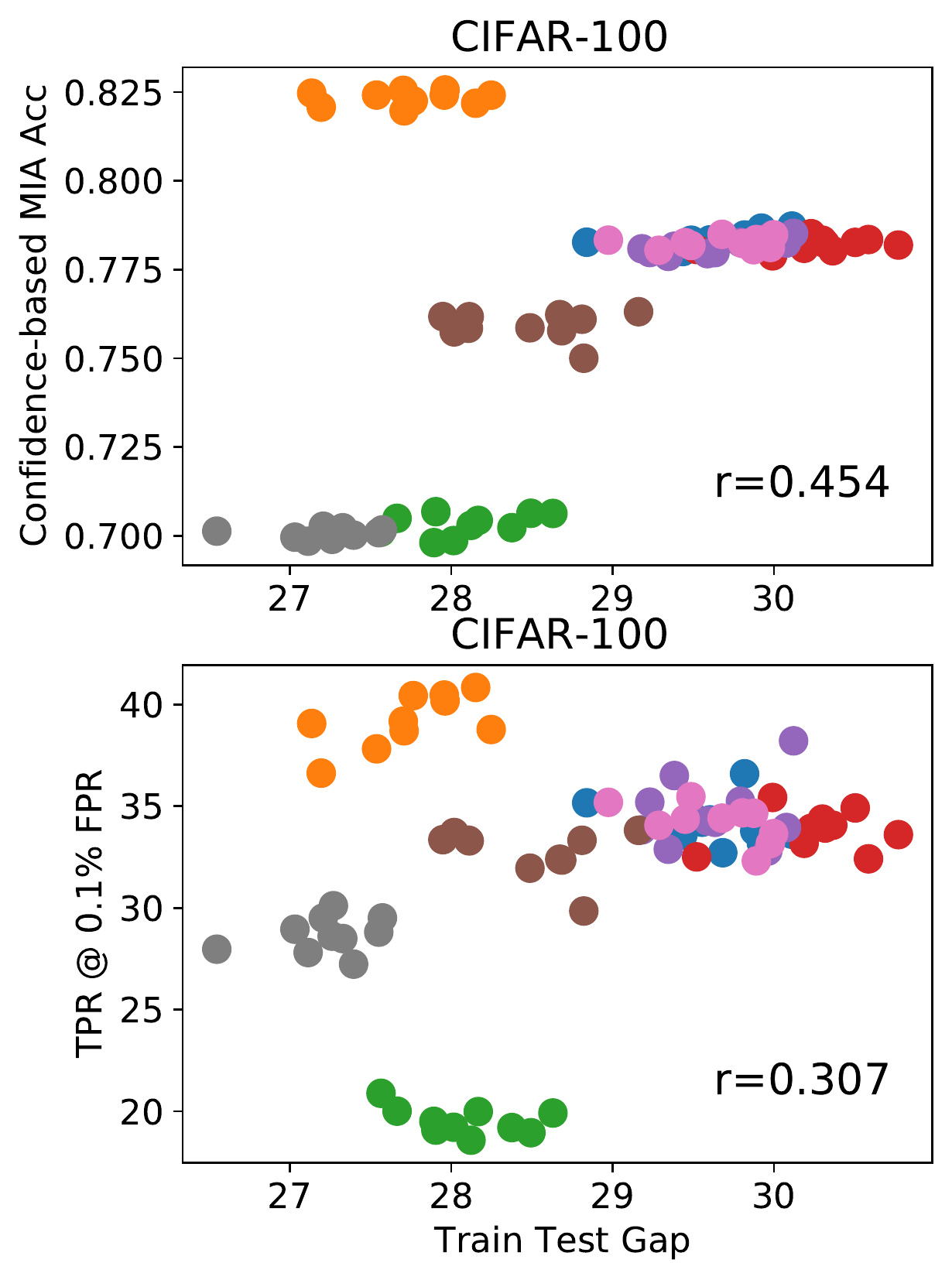}
  \caption{CIFAR-100}
\end{subfigure}
\begin{subfigure}[b]{0.385\linewidth}
  \includegraphics[height=200pt]{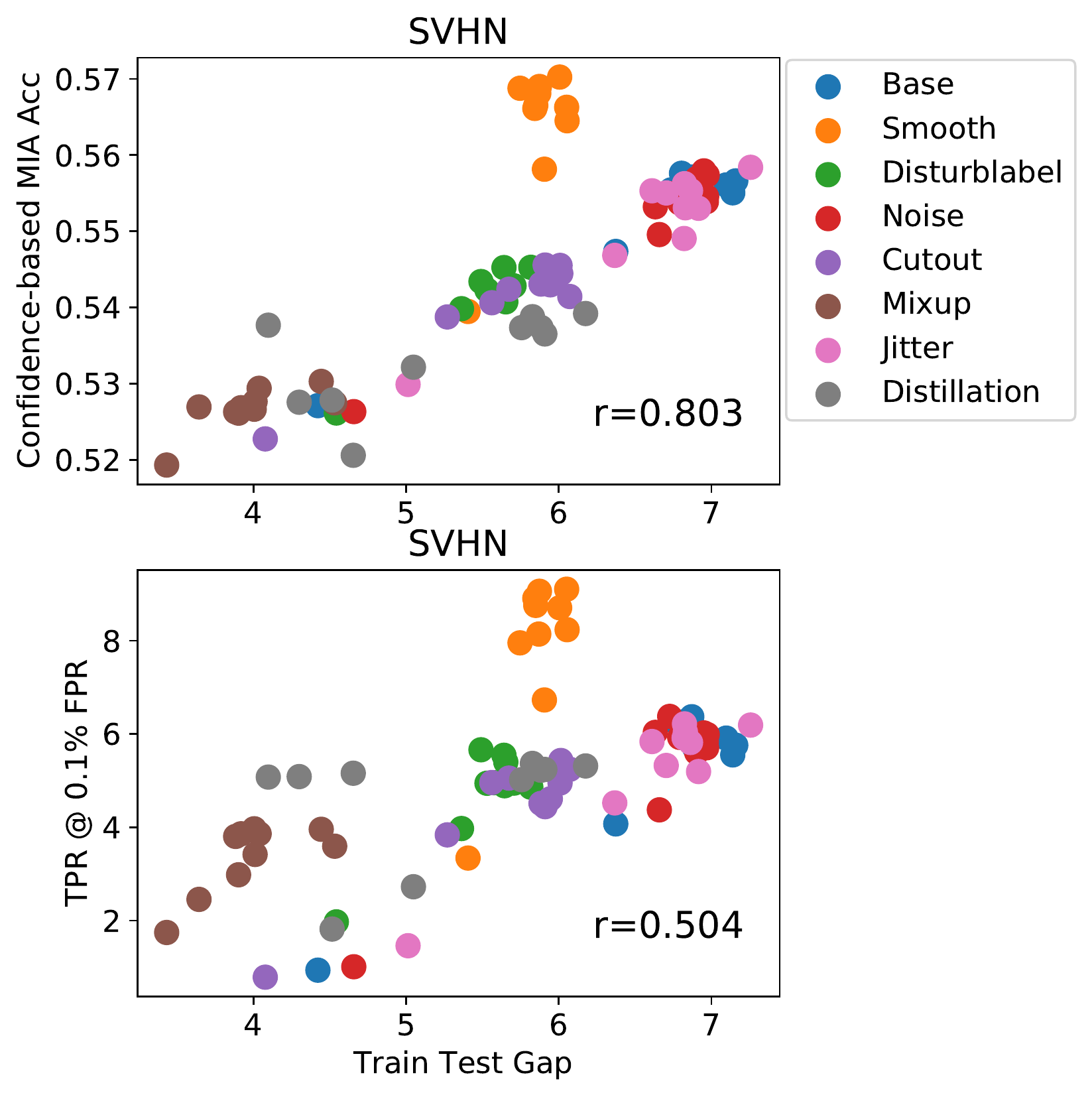}
  \caption{SVHN}
\end{subfigure}
\caption{Attack success rate versus the train-test gap of different data augmentation models on CIFAR-10, CIFAR-100, and SVHN using MaxPreCA (top) and LiRA (bottom), respectively. $r$ stands for the Pearson correlation coefficient.}
\label{fig.tprgap}
\end{figure*}

\begin{figure}[!t]
  \centering
  \includegraphics[width=\linewidth]{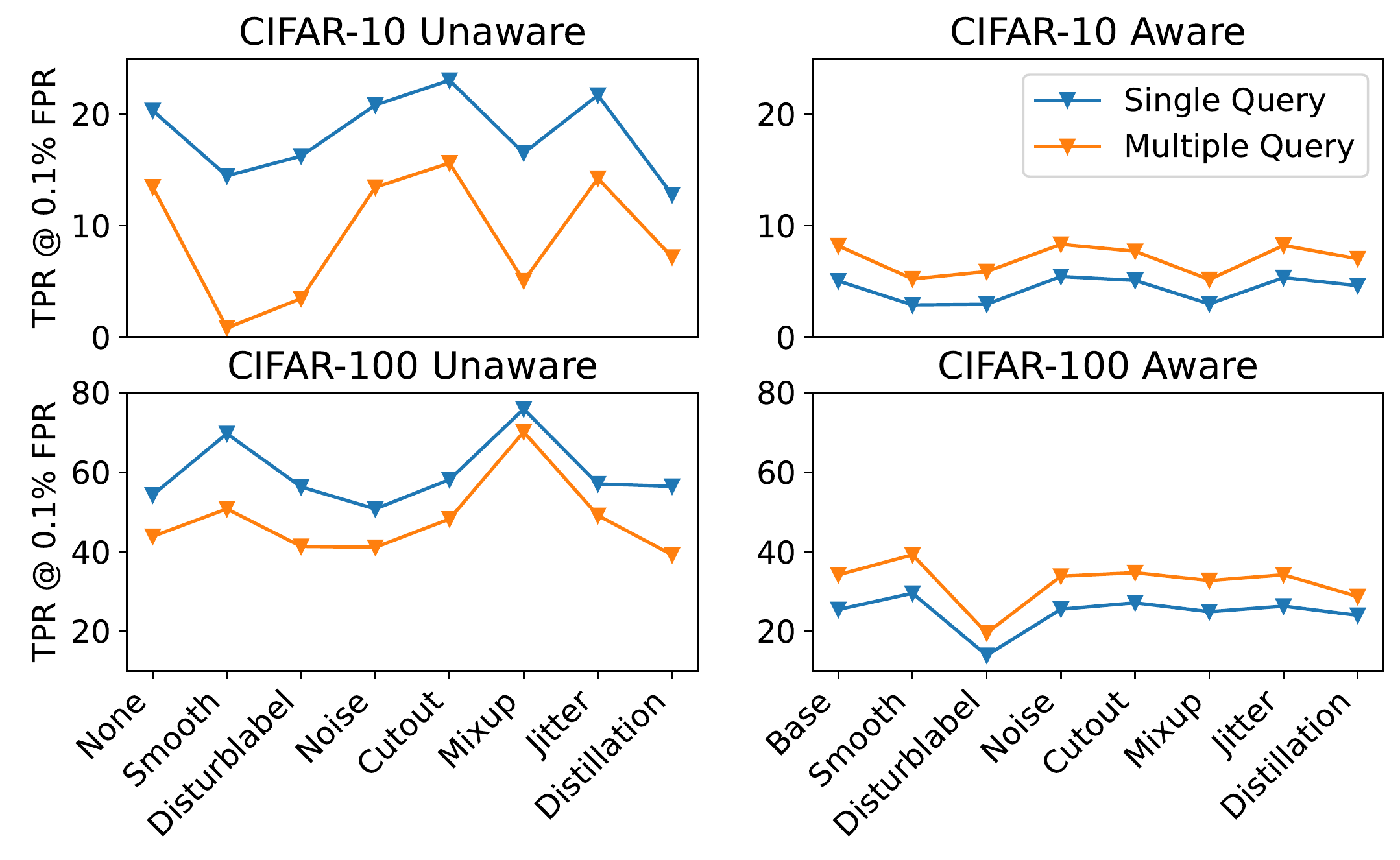}
  \caption{Attack success rates of a single query and multiple queries in two cases: augmentation-unaware (left) and augmentation-aware (right). We evaluated different data augmentation methods on CIFAR-10 and CIFAR-100 datasets, respectively. \textit{None} stands for models trained without any data augmentation.}
\label{fig.query}
\end{figure}

\begin{figure*}[!t]
  \centering
  \includegraphics[width=0.75\linewidth]{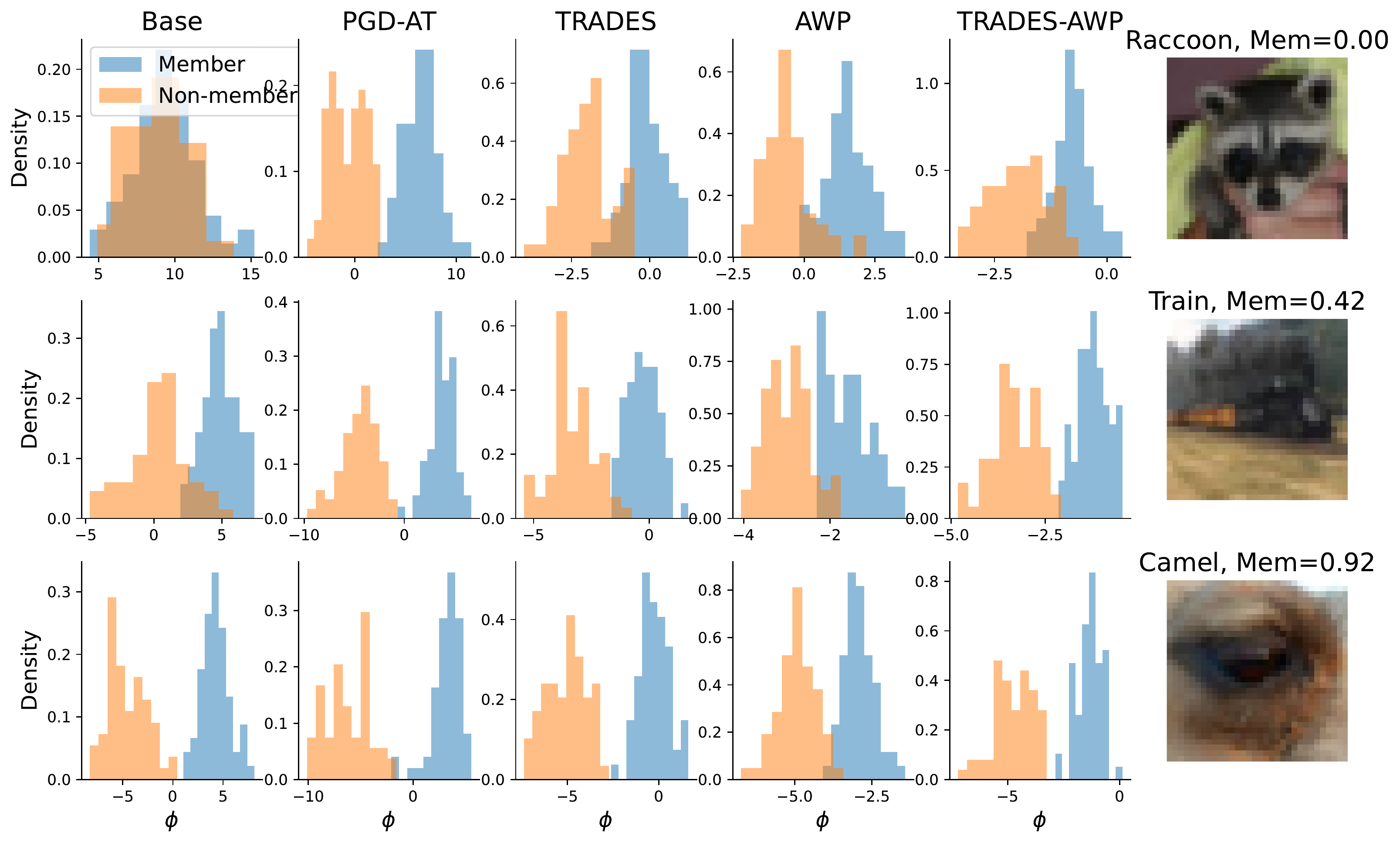}
  \caption{The distributions of normalized confidence $\phi$ of three samples with different memorization scores using Base and four adversarial training models. Each row corresponds to a sample.}
\label{fig.atEx}
\end{figure*}

\begin{figure*}[!t]
  \centering
  \includegraphics[width=0.75\linewidth]{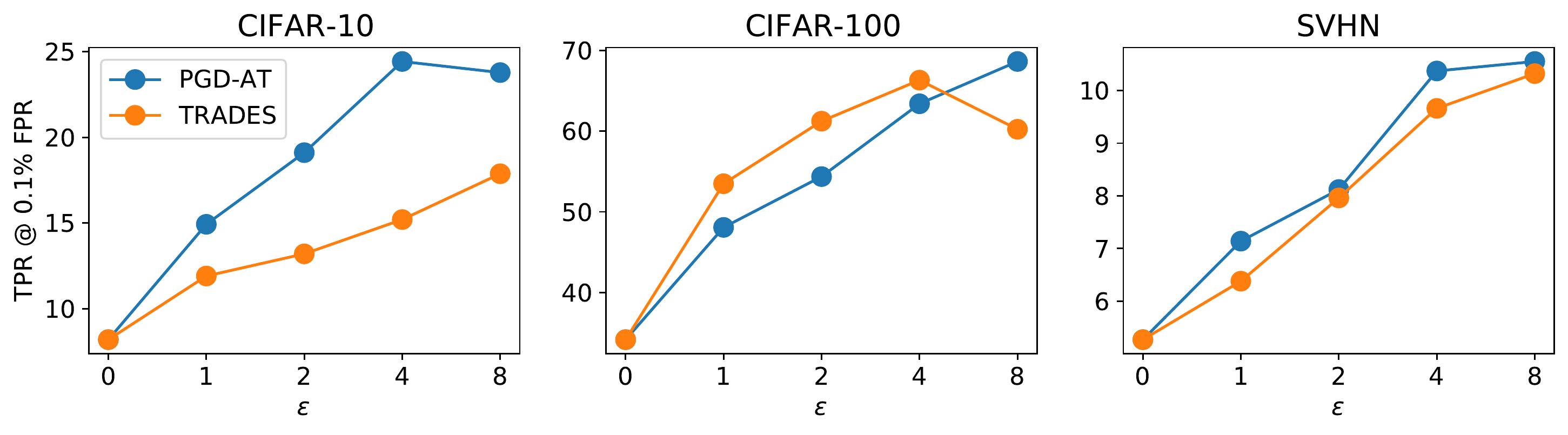}
  \caption{Attack success rates of PGD-AT and TRADES under different $\epsilon$ on the three datasets.}
\label{fig.epstpr}
\end{figure*}

\section{Privacy and Generalization Gap}
\label{sec:pri_gap}
In what follows, we proceed to analyze the relationship between privacy leakage and generalization gap.

\subsection{Compared to prior works, our results demonstrate a much weaker correlation between the privacy leakage and generalization gap.}
Many previous studies have shown that the attack success rates of MIAs are highly correlated with the generalization gap, i.e., the degree of overfitting \cite{shokri2017membership, salem2018ml, song2019privacy, yeom2018privacy, leino2020stolen}.  
To verify whether such a high correlation is still true from the memorization perspective,  in \cref{fig.tprgap}, we demonstrate the attack success rate in terms of the train-test accuracy gap of all data augmentation models for all three datasets using MaxPreCA and LiRA, respectively. 
It is obvious that compared to the MaxPreCA attack results, our results demonstrate a more scattered distribution. We also compute the Pearson correlation coefficient $r$ for each plot. As shown there, the Pearson correlation coefficients $r$ of our results are significantly lower than the results using MaxPreCA, e.g., for CIFAR-10, our $r$ is only 0.221, which is much lower than 0.708 using MaxPreCA. Note that the results using other MIAs in \cref{subsec.ave} are similar to MaxPreCA (see \cref{fig.epstpr_loss} in  Appendix \ref{supp:tau}).
Hence, via the lens of memorization, the generalization gap and privacy leakage appear less correlated than those of the previous results.

It is easy to understand why the results of these attacks are sensitive to the generalization gap, as their success rate depends heavily on how different the model behaves for training and test samples. We remark that there is a distinction between memorization and overfitting:  memorization is only necessary but not sufficient for overfitting  \cite{feldman2020does}, i.e., memorizing some training samples does not always cause overfitting. In fact, it has been both theoretically proved and empirically verified in previous work \cite{feldman2020does, feldman2020neural} that memorizing certain long-tailed samples will help in decreasing the generalization gap.  As a consequence, many attacks deployed in previous work might underestimate privacy leakage for non-overfitted models, making the correlation coefficient unnecessarily high. This issue can be alleviated in our setting as we measure the privacy leakage via the memorization perspective, the root cause of privacy leakage.

We remark that even though \citet{yeom2018privacy} also pointed out that overfitting is not the only reason for causing vulnerability to privacy attacks, they did not explicitly identify what are other factors and their attack results still demonstrate a higher correlation compared with ours (as shown in \cref{fig.tprgap}). A related observation is made in  Sections V-E of LiRA \cite{carlini2021membership} showing that there are cases where two models have similar generalization gaps but the privacy leakages vary a lot.  However, it only gives one counterexample and does not investigate it deeply. Hence, their results can only support the claim that reducing the generalization gap does not necessarily make the model less vulnerable to privacy attacks, while we take a step further and demonstrate stronger results showing that via the memorization perspective, the correlation between privacy leakage and the generalization gap becomes weaker. To the best of our knowledge, such results have not been reported in existing work.  

\subsection{Data augmentation is not necessarily an effective defense for MIAs.}
By inspecting the attack results for models with data augmentation, we can see that the privacy effects vary significantly across different data augmentation methods. For example, Distillation and Disturblabel are shown effective in reducing the vulnerability to privacy attack. Mixup, Cutout, Jitter, and Gaussian noise methods do not seem to have big impacts on the attack success rate. The main reason is that applying data augmentation does not always reduce the memorization scores of training samples, e.g., the Jitter model shown in \cref{fig.memCons}.

Moreover, among all data augmentation methods, label smoothing has drawn attention as it has been shown in both \citet{kaya2021does} and \citet{hintersdorftrust} that applying label smoothing will make the model more susceptible to MIAs. To verify this, we computed the balanced accuracy using  MaxPreCA  \cite{salem2018ml}. As shown in the top panels of \cref{fig.tprgap}, label smoothing does increase the attack accuracies compared to Base for all datasets. However, from the memorization perspective, it did not demonstrate the same tendency. By inspecting the bottom panels of \cref{fig.tprgap} we note that label smoothing demonstrates an inconsistent behavior on different datasets. On CIFAR-100 and SVHN the privacy leakage is higher than Base while on CIAFR-10 the privacy leakage is lower.   Hence, the claim that label smoothing would consistently amplify privacy leakage is untrue. Overall, we conclude that it is difficult to give a general claim about whether data augmentation can help mitigate the privacy attack or not. We remind that extra attention should be paid when relying on data augmentation as a defense technique against MIAs.

\subsection{Multiple queries can only enhance the attack if the augmentation method is known.}
As stated in \cref{sec:intro}, previous studies have shown that using augmented data to conduct multiple queries would enhance the attack success rate. To investigate this, we queried the target model using ten augmented counterparts generated by the Base method for each data point. We then targeted all the data augmentation models trained on Base as the augmentation-aware case. The augmentation-unaware case was then evaluated by targeting the data augmentation models without using Base (such models were trained on CIFAR-10 and CIFAR-100, see Appendix \ref{supp:b}). As shown in \cref{fig.query}, multiple queries did help improve the attack success rate for the augmentation-aware case, whereas, for the augmentation-unaware case, they resulted in an opposite effect, i.e., lowering the attack success rate. We note that similar results have been reported in \citet{carlini2021membership} but only with Base augmentation.  Our results further confirm their results by extending to several different data augmentation methods.

\begin{figure*}[!t]
  \centering
  \includegraphics[width=0.75\linewidth]{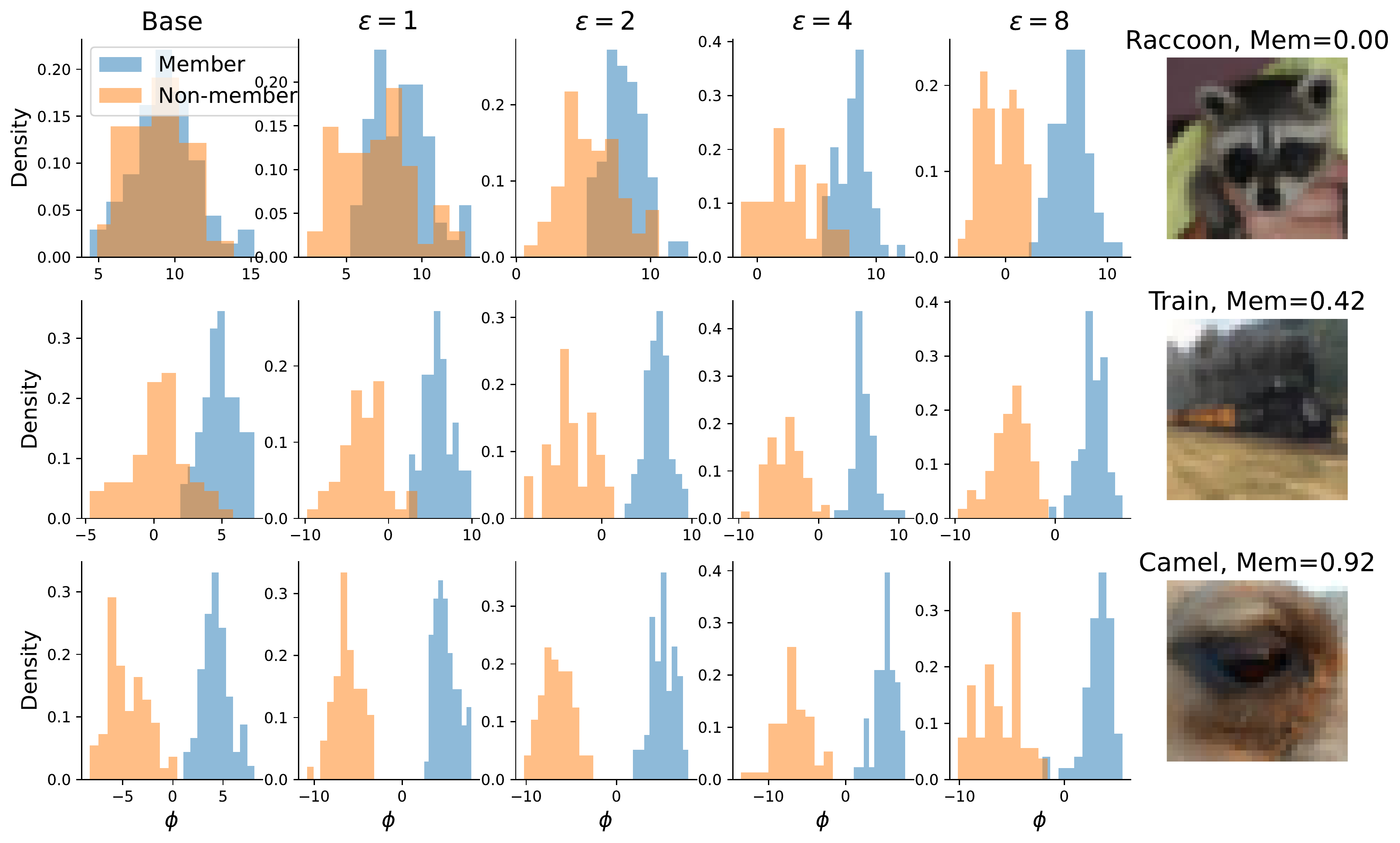}
  \caption{The distributions of normalized confidence $\phi$ of three samples with different memorization scores using Base and PGD-AT under four different $\epsilon$. Each row corresponds to a sample.}
\label{fig.epsEx}
\end{figure*}

\section{Privacy and Adversarial Robustness}
\label{sec:pri_adv}
In this section, we further analyze the relationship between privacy leakage and adversarial robustness.

\subsection{Applying adversarial training will make the model memorize more training samples, thereby causing more privacy leakages compared to the standardly trained models.}
As shown in Tables \ref{tab:c10}, \ref{tab:c100}, and \ref{tab:svhn}, applying adversarial training significantly increases privacy leakage compared to the standardly trained models. For example, the TPR @ 0.1\% FPR increases from 34.17\% to 60.23\% for Base to TRADES on CIFAR-100. One reason is that applying adversarial training will force the model to fit all the adversarial examples found in the $\ell_{\infty}$ ball around each training sample, which often increases the influence of each sample on the trained model, thereby resulting in a higher privacy risk. To visualize the effect of applying adversarial training, in \cref{fig.atEx}, we choose three examples in CIFAR-100 with different memorization scores and draw their corresponding distributions of normalized confidence $\phi$ evaluated by IN and OUT models using Base and all four types of adversarially trained models. Clearly, if the IN and OUT distributions of a particular sample are more separated, it implies that the sample is adversarial training a higher privacy risk.  We can see that for samples that have low privacy risks (e.g., \textit{Raccoon} and \textit{Train}), applying adversarial training would make the distribution more separable.  

Note that there is a bottleneck of increasing the privacy risk of samples when performing adversarial training: for samples that are already at a high privacy risk (e.g., \textit{Camel} with a high memorization score), applying adversarial training would not make much difference as the distributions are already quite separated for standardly trained models. Overall, we conclude that one major reason why adversarial training causes a higher privacy leakage is that it memorizes many training samples that are not memorized by standardly trained models.

\begin{table}[!t]
  \centering
  \small
	\renewcommand{\arraystretch}{1.15}{
  \setlength{\tabcolsep}{4pt}
  {
    \begin{tabular}{c|c|cc}
	\multirow{2}*{\textbf{Dataset}} & \multirow{2}*{\textbf{Method}} & \textbf{Adversarial} & \textbf{TPR @}  \\
	 &  & \textbf{Acc} & \textbf{0.1\% FPR} \\
	\hline
	\multirow{5}*{CIFAR-10} 
        & Base & $0.0 \pm 0.0$ & $8.20 \pm 0.45$ \\
        \cline{2-4}
        & PGD-AT & $38.8 \pm 0.4$ & $23.78 \pm 0.89$ \\
	& TRADES & $45.2 \pm 0.3$ & $17.88 \pm 1.56$ \\
	& AWP & $45.9 \pm 0.1$ & $10.58 \pm 3.48$ \\
	& TRADES-AWP & $48.8 \pm 0.2$ & $12.43 \pm 0.89$ \\
	
     \hline
    \multirow{5}*{CIFAR-100} 
        & Base & $0.0 \pm 0.0$ & $34.17 \pm 1.05$ \\
        \cline{2-4}
        & PGD-AT & $16.9 \pm 0.1$ & $68.63 \pm 0.88$ \\
	& TRADES & $19.7 \pm 0.4$ & $60.23 \pm 0.87$ \\
	& AWP & $23.9 \pm 0.1$ & $39.92 \pm 2.40$ \\
	& TRADES-AWP & $23.3 \pm 0.2$ & $57.51 \pm 2.02$ \\
    \hline
    \multirow{5}*{SVHN}
    & Base & $0.0 \pm 0.0$ & $5.27 \pm 1.57$ \\
    \cline{2-4}
        & PGD-AT & $45.2 \pm 0.4$ & $10.55 \pm 0.25$ \\
	& TRADES & $47.4 \pm 0.1$ & $10.31 \pm 0.22$ \\
	& AWP & $43.7 \pm 0.5$ & $8.85 \pm 0.24$ \\
	& TRADES-AWP & $46.2 \pm 0.4$ & $8.54 \pm 0.22$ \\
    \end{tabular}
    
    }
   }
   \caption{The accuracies on adversarial examples (Adversarial Acc) and privacy leakage of different adversarial training models on three datasets. The accuracies are evaluated using PGD with $\epsilon = 8$ and 20 iteration steps.}
  \label{tab:robustness}
\end{table}

\subsection{Better adversarial robustness does not necessarily make the adversarially trained model more vulnerable to privacy attacks.} To further investigate the relation between adversarial robustness and privacy leakage, in \cref{tab:robustness} we compare adversarial robustness and attack success rate using different adversarial training methods on the three datasets. We can see that compared to TRADES and PGD-AT,  both AWP and TRADES-AWP achieve higher adversarial accuracies, while the attack success rates are lower. Hence, for adversarially trained models, stronger adversarial robustness does not necessarily come with a cost of
privacy leakage.


In addition to the attack results using different adversarial training methods, it is also interesting to see how the attack result changes along with varying parameters. Since $\epsilon$ is a critical parameter for adversarial training, in \cref{fig.epstpr}, we compare the attack results of different $\epsilon$ using both PGD-AT and TRADES models on the three datasets. We can see that overall the attack success rate tends to increase along with $\epsilon$ (at least for $\epsilon<8$). One reason might be that increasing the perturbation parameter $\epsilon$ will result in a bigger $\ell_{\infty}$ ball around each training sample thereby increasing the difficulty of fitting all adversarial examples. Consequently, the model will be more sensitive to this sample, thus increasing the privacy risk.  To verify this,  in \cref{fig.epsEx}, we use three examples (the same as in \cref{fig.atEx}) with different memorization scores and draw their corresponding distributions of normalized confidence $\phi$ evaluated by IN and OUT models using Base and PGD-AT models under four different $\epsilon$. Clearly, we can see that for the sample with a very low memorization score (\textit{Raccon}), increasing $\epsilon$ makes the distributions more separable thus posing a higher privacy risk. Similar to \cref{fig.atEx}, we observe a bottleneck effect that for samples with high memorization score (\textit{Camel}), varying $\epsilon$ will not make much difference.  Moreover, such bottleneck effect is quicker since for samples with a fair memorization score (\textit{Train}), the IN and OUT distributions are already quite separable when $\epsilon=1$.   

\section{Discussions}
\label{sec:disscussions}
This section extends our discussion to encompass a broader range of experimental results, providing further evidence of the generalizability of our findings across diverse model architectures, datasets, and MIAs.

\subsection{Expanded Experiments across Diverse Model Architectures and Datasets}

To further verify the generalizability of our key conclusions, initially derived from experiments on three image datasets (CIFAR-10, CIFAR-100, and SVHN) using ResNet-18, ResNet-18, and CNN-8, respectively, we conducted additional experiments involving other combinations of datasets and architectures. Specifically, we replicated the experiments using CNN-8 on CIFAR10 as an alternative combination. Furthermore, we performed similar experiments with an MLP architecture on two additional non-image datasets, namely Purchases and Locations processed by \citet{shokri2017membership}.

\begin{figure}[!t]
  \centering
  \begin{subfigure}[b]{0.42\linewidth}
      \includegraphics[height=150pt]{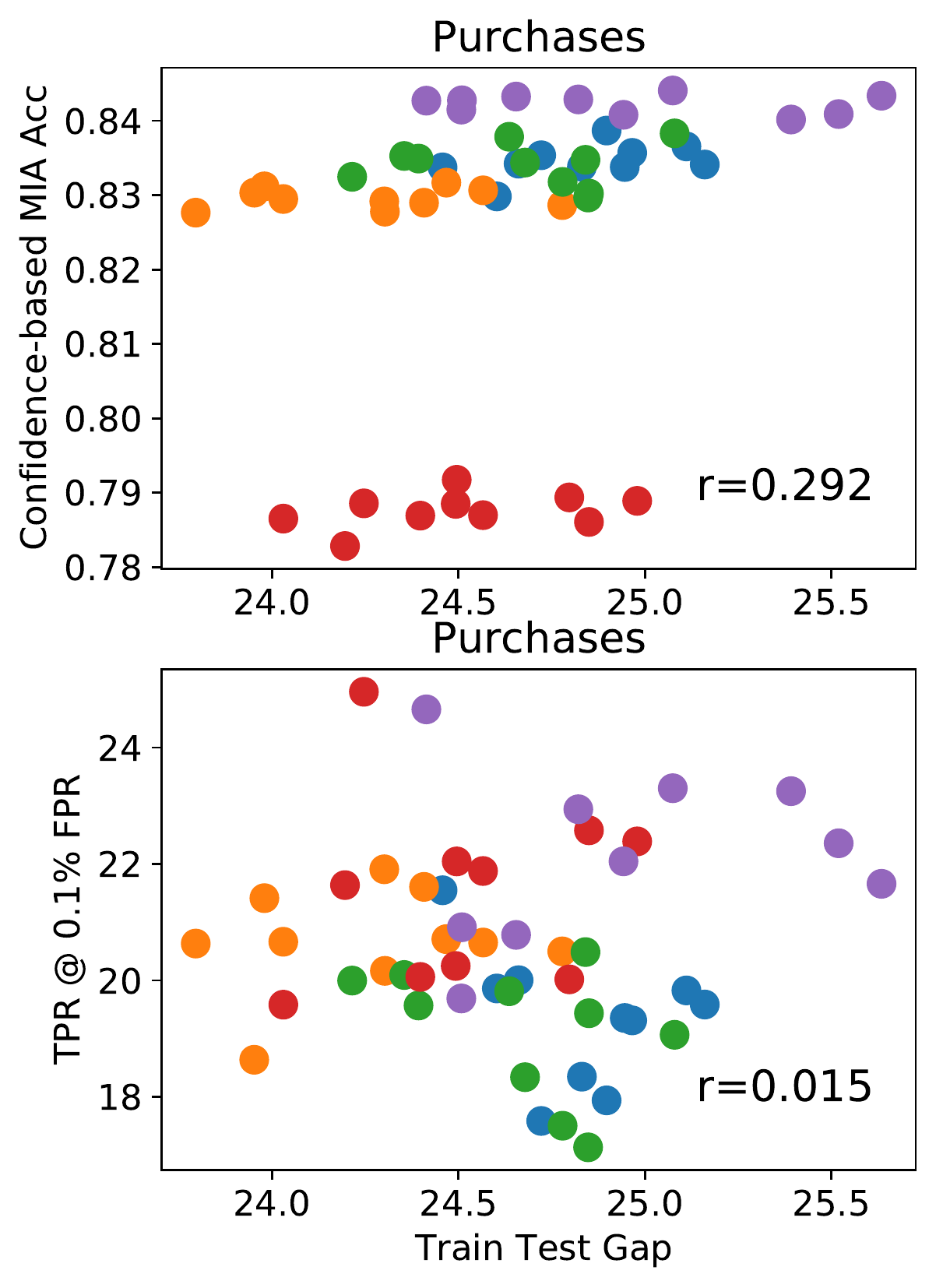}
      \caption{Purchases}
  \end{subfigure}
\begin{subfigure}[b]{0.56\linewidth}
  \includegraphics[height=150pt]{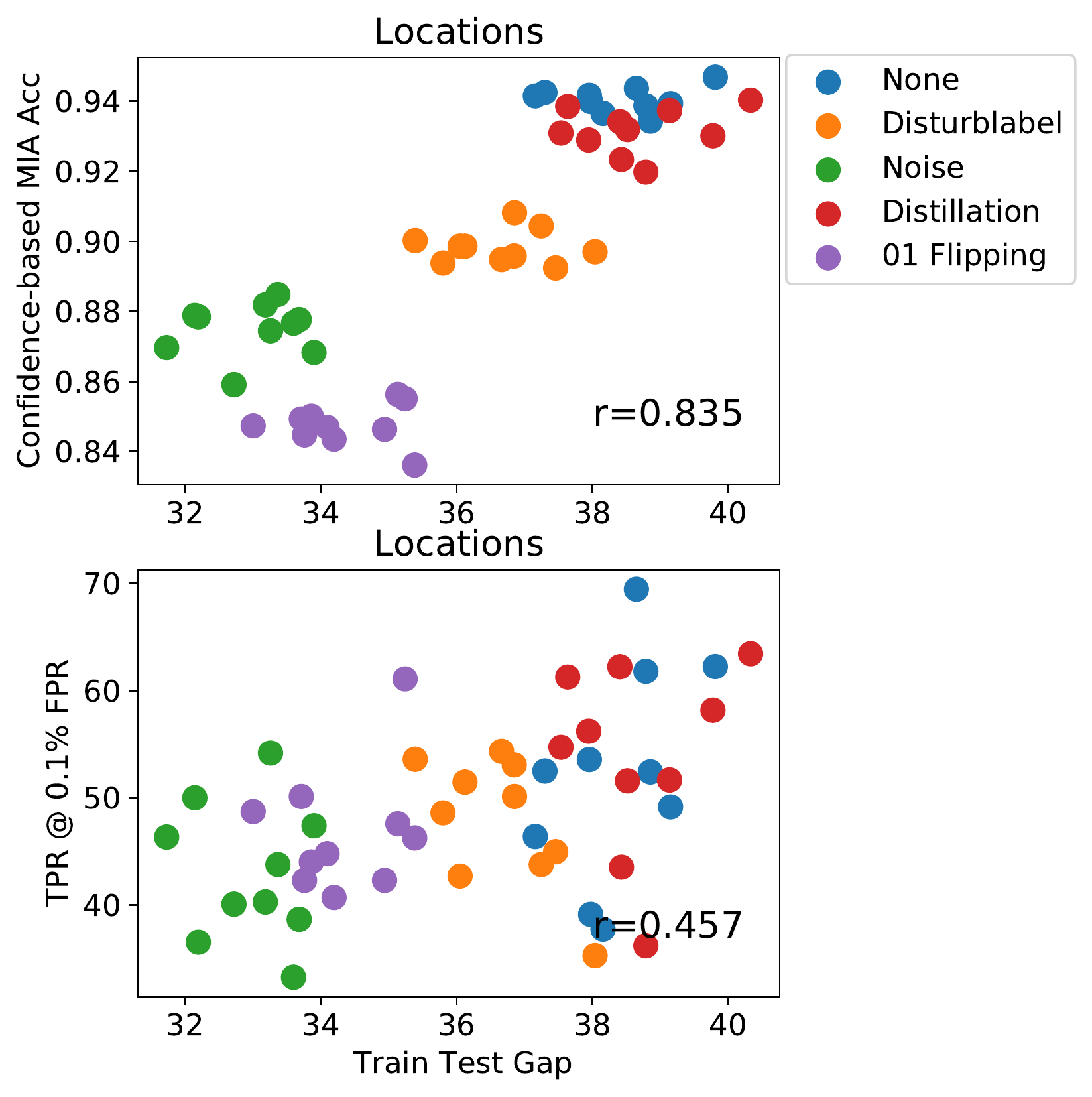}
  \caption{Locations}
\end{subfigure}
  \caption{Attack success rate versus the train-test gap of different data augmentation models on the Purchases and Locations datasets using MaxPreCA (top) and LiRA (bottom), respectively. $r$ stands for the Pearson correlation coefficient.}
\label{fig.epstpr_locations}
\end{figure}

In Appendix \ref{supp:cnn}, we present the results of experiments conducted using CNN-8 on CIFAR-10. These findings consistently align with our previous key observations: 1) A less pronounced correlation between generalization gaps and privacy leakages, as indicated by the reduced correlation coefficient in \cref{fig.epstpr_cifar10_cnn}; 2) Adversarially trained models memorize more training data compared to standardly trained models, as evidenced by the much higher privacy leakage in \cref{tab:robustness_cifar10_cnn}; 3) An increase in the adversarial robustness of the adversarially trained models does not necessarily come with a price of heightened vulnerability to privacy attacks, as demonstrated by the performance of AWP in \cref{tab:robustness_cifar10_cnn}. Moreover, the trends depicted in \cref{fig.epstpr_cnn} closely resemble those observed with the ResNet-18 architecture on CIFAR-10 in \cref{fig.epstpr}.

For the non-image datasets, we performed experiments with an MLP architecture on the Purchases and Locations datasets. We note that part of the evaluated data augmentation methods relevant to the image datasets, e.g., Cropping, Jitter, Cutout, etc, are inapplicable to these non-image datasets. Here we investigated the privacy impacts of applying Disturblabel, Distillation, and Gaussian Noise. Additionally, we investigated a 0-1 Flipping augmentation method specially designed for these datasets. We did not evaluate adversarial training methods with the non-image datasets because adversarial training is primarily focused on the image domain \cite{madry2018towards, zhang2019theoretically, awp}, and there is no standardized definition of adversarial examples for these non-image datasets. The privacy leakage results and the experimental details are presented in Appendix \ref{supp:nonImg}. In \cref{fig.epstpr_locations}, we demonstrate the attack success rate regarding the train-test accuracy gap of all data augmentation models for these non-image datasets using MaxPreCA and LiRA, respectively. We can see that our results with LiRA demonstrate a more dispersed distribution and a lower Pearson correlation coefficient compared with the MaxPreCA attack results. 

Overall, we conclude that our key findings remain consistent across diverse model architectures and datasets.

\subsection{Other MIAs Beyond LiRA}
\label{sec:beyond}
While our initial investigations of privacy leakage focused on LiRA, it is important to clarify that our primary goal was to assess the privacy implications of data enhancement in machine learning models, particularly via the perspective of memorization.  The choice of LiRA was driven by its high consistency with memorization scores, as demonstrated in \cref{fig.mem}. However, our findings are not limited to LiRA alone; they extend to other MIAs that demonstrate a strong correlation with memorization scores. To validate this,  we shifted our attention to the difficulty calibrated loss approach \cite{importance}, which shows comparable consistency with memorization scores (see \cref{fig.mem}(e)). The results, depicted in \cref{fig.epstpr_tausingle} and \cref{tab:robustness_newattack}, indicate that the general trends in privacy effects resulting from data augmentation and adversarial training align with those observed using LiRA (see \cref{fig.epstpr} and \cref{tab:robustness}).  It is worth noting that the correlation score is slightly higher and the overall attack performance is lower compared to LiRA, which aligns with the expectations given LiRA's higher consistency and superior attack performance against the importance calibrated loss approach. Consequently, we conclude that our findings apply to other MIAs as long as they exhibit a high level of consistency with memorization scores.

\section{Conclusion} 
\label{sec:conclu}

In this paper, we reinvestigate the privacy effect of applying data augmentation and adversarial training to machine learning models via a new perspective, namely the degree of memorization. Such reinvestigation is quite necessary as we found that the attacks deployed in previous studies for measuring privacy leakage produce misleading results: the training samples with low privacy risks are more prone to be identified as members compared to the ones with high privacy risks. Through a systematic evaluation, we reveal some findings conflict with previous results, e.g., the generalization gap and privacy leakage are shown less correlated than those of the previous results and label smoothing does not always amplify the privacy leakage.  Moreover, we also show that improving the adversarial robustness (via adversarial training) does not necessarily make the adversarially trained model more vulnerable to privacy attacks. Our results call for more investigations on the privacy of machine learning models from the memorization perspective.

\section*{Acknowledgments}
This work was supported by the National Natural Science Foundation of China (Nos. U2341228 and U19B2034).

\bibliographystyle{IEEEtranN}
\bibliography{egbib.bib}

\clearpage
\appendices
\setcounter{table}{0}
\renewcommand{\thetable}{B.\arabic{table}}

\begin{table*}[!t]
  \centering
  \small
\renewcommand{\arraystretch}{1.15}{
  \setlength{\tabcolsep}{4pt}
  {
    \begin{tabular}{c|cc|cccc}
    \textbf{Method} & \textbf{Training Acc}  & \textbf{Test Acc} & \textbf{TPR @ 0.1\% FPR} & \textbf{TPR @ 0.001\% FPR} & \textbf{Log-scale AUC} & \textbf{MIA Balanced Acc} \\
    
     \hline
Base & $99.2 \pm 0.8$ & $92.6 \pm 0.2$     & $5.27 \pm 1.57$ & $2.06 \pm 0.85$     & $0.778 \pm 0.040$     & $57.71 \pm 1.19$ \\
\hline
Smooth & $99.5 \pm 0.3$ & $93.7 \pm 0.2$     & $\bm{7.89} \pm 1.66$ & $\bm{2.89} \pm 1.22$     & $\bm{0.809} \pm 0.022$     & $\bm{60.34} \pm 1.27$ \\
Disturblabel & $98.7 \pm 0.5$ & $93.2 \pm 0.2$     & $4.72 \pm 1.02$ & $1.64 \pm 0.83$     & $0.771 \pm 0.028$     & $57.26 \pm 0.65$ \\
Noise & $99.2 \pm 0.9$ & $92.6 \pm 0.3$     & $\bm{5.29} \pm 1.51$ & $\bm{2.17} \pm 0.93$     & $\bm{0.780} \pm 0.037$     & $57.68 \pm 1.23$ \\
Cutout & $98.7 \pm 0.8$ & $93.1 \pm 0.3$     & $4.38 \pm 1.27$ & $1.66 \pm 0.83$     & $0.767 \pm 0.037$     & $57.02 \pm 1.10$ \\
Mixup & $97.4 \pm 0.5$ & $93.4 \pm 0.3$     & $3.37 \pm 0.71$ & $0.89 \pm 0.42$     & $0.749 \pm 0.020$     & $56.43 \pm 0.66$ \\
Jitter & $99.2 \pm 0.6$ & $92.6 \pm 0.2$     & $5.25 \pm 1.36$ & $\bm{2.17} \pm 0.94$     & $\bm{0.781} \pm 0.029$     & $57.59 \pm 1.00$ \\
Distillation & $95.1 \pm 5.4$ & $90.8 \pm 3.4$     & $3.10 \pm 2.27$ & $1.29 \pm 1.11$     & $0.691 \pm 0.120$     & $54.71 \pm 2.69$ \\
\hline
PGD-AT & $97.9 \pm 0.5$ & $86.2 \pm 0.3$     & $\bm{10.55} \pm 0.25$ & $\bm{4.31} \pm 0.13$     & $\bm{0.852} \pm 0.011$     & $\bm{65.23} \pm 0.49$ \\
TRADES & $94.3 \pm 0.7$ & $82.5 \pm 0.8$     & $\bm{10.31} \pm 0.22$ & $\bm{4.22} \pm 0.16$     & $\bm{0.837} \pm 0.014$     & $\bm{65.09} \pm 0.74$ \\
AWP & $96.5 \pm 0.7$ & $85.2 \pm 0.6$     & $\bm{8.85} \pm 0.24$ & $\bm{3.08} \pm 0.05$     & $\bm{0.820} \pm 0.020$     & $\bm{62.85} \pm 0.59$ \\
TRADES-AWP & $91.4 \pm 1.4$ & $80.2 \pm 1.2$     & $\bm{8.54} \pm 0.22$ & $\bm{3.05} \pm 0.04$     & $\bm{0.794} \pm 0.042$     & $\bm{62.76} \pm 0.63$ \\
    \end{tabular}
    }
    }
  \caption{Attack success rates of different data enhancement on SVHN. The same conventions are used as in \cref{tab:c10}. }
  \label{tab:svhn}
\end{table*}

\begin{table*}[!t]
  \centering
  \small
\renewcommand{\arraystretch}{1.15}{
  \setlength{\tabcolsep}{4pt}
  {
    \begin{tabular}{c|cc|cccc}
    \textbf{Method} & \textbf{Training Acc}  & \textbf{Test Acc} & \textbf{TPR @ 0.1\% FPR} & \textbf{TPR @ 0.001\% FPR} & \textbf{Log-scale AUC} & \textbf{MIA Balanced Acc} \\
    
     \hline
None & $100.0 \pm 0.0$ & $82.9 \pm 0.5$     & $20.35 \pm 4.31$ & $9.44 \pm 3.24$     & $0.885 \pm 0.013$     & $76.25 \pm 2.18$ \\
\hline
None + Smooth & $100.0 \pm 0.0$ & $83.7 \pm 0.5$     & $14.48 \pm 3.03$ & $2.37 \pm 1.79$     & $0.839 \pm 0.024$     & $72.91 \pm 1.60$ \\
None + Disturblabel & $100.0 \pm 0.0$ & $84.1 \pm 0.6$     & $16.26 \pm 1.03$ & $3.43 \pm 2.48$     & $0.853 \pm 0.016$     & $72.34 \pm 0.89$ \\
None + Noise & $100.0 \pm 0.0$ & $82.4 \pm 0.7$     & $\bm{20.84} \pm 3.69$ & $8.59 \pm 2.88$     & $\bm{0.886} \pm 0.011$     & $\bm{76.90} \pm 2.52$ \\
None + Cutout & $100.0 \pm 0.0$ & $84.0 \pm 0.6$     & $\bm{23.07} \pm 0.80$ & $\bm{10.53} \pm 1.85$     & $\bm{0.894} \pm 0.004$     & $\bm{77.42} \pm 0.37$ \\
None + Mixup & $100.0 \pm 0.0$ & $83.7 \pm 0.4$     & $16.53 \pm 1.42$ & $4.84 \pm 1.58$     & $\bm{0.867} \pm 0.006$     & $\bm{76.53} \pm 0.99$ \\
None + Jitter & $100.0 \pm 0.0$ & $82.0 \pm 1.3$     & $\bm{21.73} \pm 5.85$ & $8.77 \pm 3.64$     & $\bm{0.886} \pm 0.020$     & $\bm{77.63} \pm 2.61$ \\
None + Distillation & $100.0 \pm 0.0$ & $84.4 \pm 0.4$     & $12.79 \pm 1.81$ & $5.16 \pm 1.18$     & $0.851 \pm 0.010$     & $69.25 \pm 0.99$ \\

    \end{tabular}
    
    }
    }
  \caption{Attack success rates of different data augmentation methods \textbf{without} using Base on CIFAR-10. The 2nd and 3rd columns show the training and test accuracies of each method, respectively. The 4th - 7th columns show four metrics to evaluate the extent of privacy leakages. We highlight the MIA success rates for different data augmentation methods that are larger than that for None.}
  \label{tab:c10_none}
\end{table*}

\begin{table*}[!t]
  \centering
  \small
\renewcommand{\arraystretch}{1.15}{
  \setlength{\tabcolsep}{4pt}
  {
    \begin{tabular}{c|cc|cccc}
    \textbf{Method} & \textbf{Training Acc}  & \textbf{Test Acc} & \textbf{TPR @ 0.1\% FPR} & \textbf{TPR @ 0.001\% FPR} & \textbf{Log-scale AUC} & \textbf{MIA Balanced Acc} \\
    \hline
    None & $100.0 \pm 0.0$ & $54.0 \pm 0.9$     & $54.26 \pm 10.72$ & $30.68 \pm 11.03$     & $0.954 \pm 0.014$     & $92.96 \pm 2.28$ \\
     \hline
    None + Smooth & $100.0 \pm 0.0$ & $53.7 \pm 2.0$     & $\bm{69.71} \pm 3.51$ & $\bm{41.81} \pm 9.55$     & $\bm{0.972} \pm 0.004$     & $\bm{96.76} \pm 0.33$ \\
    None + Disturblabel & $100.0 \pm 0.0$ & $55.5 \pm 0.5$     & $\bm{56.30} \pm 1.22$ & $\bm{37.37} \pm 4.58$     & $\bm{0.959} \pm 0.003$     & $90.96 \pm 0.23$ \\
    None + Noise & $100.0 \pm 0.0$ & $53.5 \pm 1.2$     & $50.75 \pm 8.61$ & $28.61 \pm 10.75$     & $0.950 \pm 0.011$     & $91.79 \pm 1.70$ \\
    None + Cutout & $100.0 \pm 0.0$ & $54.1 \pm 0.8$     & $\bm{58.14} \pm 5.60$ & $\bm{36.64} \pm 7.94$     & $\bm{0.961} \pm 0.007$     & $92.87 \pm 1.04$ \\
    None + Mixup & $100.0 \pm 0.0$ & $49.3 \pm 0.7$     & $\bm{75.88} \pm 1.09$ & $\bm{51.04} \pm 6.80$     & $\bm{0.978} \pm 0.002$     & $\bm{96.13} \pm 0.06$ \\
    None + Jitter & $100.0 \pm 0.0$ & $53.1 \pm 0.9$     & $\bm{57.05} \pm 11.22$ & $30.21 \pm 14.71$     & $0.955 \pm 0.016$     & $\bm{93.47} \pm 2.18$ \\
    None + Distillation & $100.0 \pm 0.0$ & $57.7 \pm 1.7$     & $\bm{56.45} \pm 3.95$ & $\bm{35.61} \pm 7.31$     & $\bm{0.959} \pm 0.006$     & $90.39 \pm 0.88$ \\
    \end{tabular}
    
    }
    }
  \caption{Attack success rates of different data augmentation methods  \textbf{without} using Base on CIFAR-100. The same conventions are used as in \cref{tab:c10_none}.}
  \label{tab:c100_none}
\end{table*}

\section{The Hyper-Parameters of Each Data Augmentation}
\label{supp:a}

As stated in the paper, the hyper-parameter of each data augmentation method was set to achieve relatively high test accuracy by trying various values. Here we report the values we tried and the final values used when training 128 shadow models for each data augmentation method on CIFAR-10, CIFAR100, and SVHN:
\begin{enumerate}
\item Random Cropping and Flipping: First, the images with a resolution of 32 $\times$ 32 were padded with zeros of 4 pixels on each end. Then the padded images with the resolution of 36 $\times$ 36 were randomly cropped out to form inputs with the resolution of 32 $\times$ 32. Finally, the inputs were randomly flipped horizontally. Unless otherwise specified, all other data augmentation methods also use this as default. 

\item Label smoothing: We tried $\epsilon$ including 0.01, 0.05, 0.1, 0.2, 0.3, 0.4, 0.5, 0.6, 0.7, and 0.8. Finally, we chose 0.2 on CIFAR-10, 0.3 on CIFAR-100, and 0.2 on SVHN.

\item Disturblabel: We tried $\epsilon$ including 0.01, 0.05, 0.1, 0.2, 0.3, 0.4, 0.425, 0.45, 0.5, 0.525, 0.55, 0.575, and 0.6. Finally, we chose 0.05 on CIFAR-10, 0.3 on CIFAR-100, and 0.05 on SVHN.

\item Gaussian Noise: We tried $\sigma$ including 0.025, 0.01, 0.05, 0.075, 0.1, 0.125, 0.15, 0.175, 0.2, 0.225, 0.25, 0.275, 0.3, 0.325, and 0.35. Finally, we chose $\sigma$ to be 0.01 on the three datasets.

\item Cutout: We tried $M$ including 4, 8, 12, 16, and 20. Finally, we chose $M$ to be 8 on the three datasets.

\item Mixup: $\gamma$ used in Mixup is sampled from a \textit{beta} distribution $\gamma\sim\mathrm{Be}(\alpha, \alpha)$. We tried $\alpha$ including 0.5, 0.1, 0.25, 1, 2, 4, 8, 16, 32, 64, 128, and 256. Finally, we chose $\alpha$ to be 0.5 on the three datasets.

\item Jitter: We used the ColorJitter function in Torchvision\footnote{\url{https://pytorch.org/vision/stable/index.html}} directly. We tried the parameters corresponding to brightness, contrast, saturation, and hue including 0.05, 0.1, 0.2, 0.15, 0.25, 0.3, 0.35, 0.4, 0.45, and 0.5. Finally, we chose 0.05 on the three datasets.

\item Distillation: We tried $T$ including 1, 2, 3, 5, and 10. Finally, we chose $T$ to be 3 on the three datasets.
\end{enumerate}

\section{Additional Membership Inference Attack Results}
\label{supp:b}
\renewcommand{\thetable}{B.\arabic{table}}
\setcounter{table}{0}
 The training and test accuracies and MIA results of all data augmentation models on SVHN are shown in \cref{tab:svhn}. In addition, taking CIFAR-10 and CIFAR-100 as examples, the training and test accuracies and MIA results of all data augmentation models trained without using Base are demonstrated in Tables \ref{tab:c10_none} and \ref{tab:c100_none}.  Single query was used because it obtained higher attack success rates than multiple queries, as shown in Figure 4 in the paper. Here \textit{None} stands for models trained without any data augmentation (only the original image data). The test accuracies of models trained without using Base are much lower than that of models trained using Base.

\section{Membership Inference Attack Results on CIFAR-10 with CNN-8}
\label{supp:cnn}
\renewcommand{\thefigure}{C.\arabic{figure}}
\setcounter{figure}{0}
\renewcommand{\thetable}{C.\arabic{table}}
\setcounter{table}{0}

\begin{figure}[!t]
  \centering
  \includegraphics[width=0.8\linewidth]{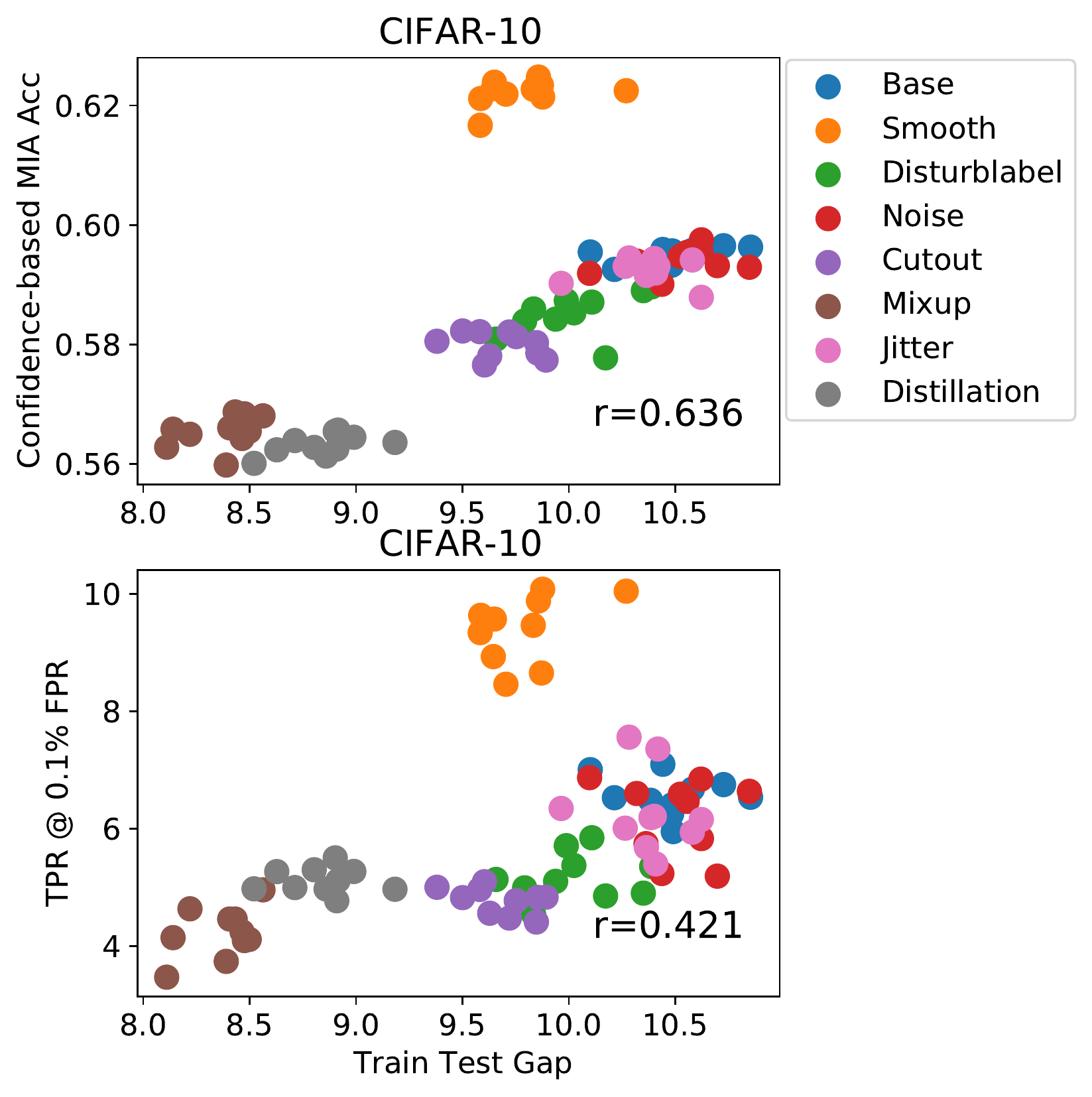}
  \caption{Attack success rate versus the train-test gap of different data augmentation models on CIFAR-10 with CNN-8 using MaxPreCA (top) and LiRA (bottom), respectively.}
\label{fig.epstpr_cifar10_cnn}
\end{figure}

\begin{table}[!t]
  \centering
  \small
	\renewcommand{\arraystretch}{1.15}{
  \setlength{\tabcolsep}{4pt}
  {
    \begin{tabular}{c|c|cc}
	\multirow{2}*{\textbf{Dataset}} & \multirow{2}*{\textbf{Method}} & \textbf{Adversarial} & \textbf{TPR @}  \\
	 &  & \textbf{Acc} & \textbf{0.1\% FPR} \\
	\hline
	\multirow{5}*{CIFAR-10} 
        & Base & $0.0 \pm 0.0$ & $6.57 \pm 0.32$ \\
        \cline{2-4}
        & PGD-AT & $36.0 \pm 0.3$ & $17.08 \pm 0.48$ \\
	& TRADES & $42.1 \pm 0.4$ & $14.25 \pm 2.23$ \\
	& AWP & $42.4 \pm 0.3$ & $9.07 \pm 1.85$ \\
	& TRADES-AWP & $43.0 \pm 0.3$ & $11.43 \pm 0.76$ \\
	
    \end{tabular}
    
    }
   }
   \caption{The accuracies on adversarial examples (Adversarial Acc) and privacy leakage of different adversarial training models on CIFAR-10 with CNN-8. The accuracies are evaluated using PGD with $\epsilon = 8$ and 20 iteration steps. }
  \label{tab:robustness_cifar10_cnn}
\end{table}

\begin{figure}[!t]
  \centering
  \includegraphics[width=0.65\linewidth]{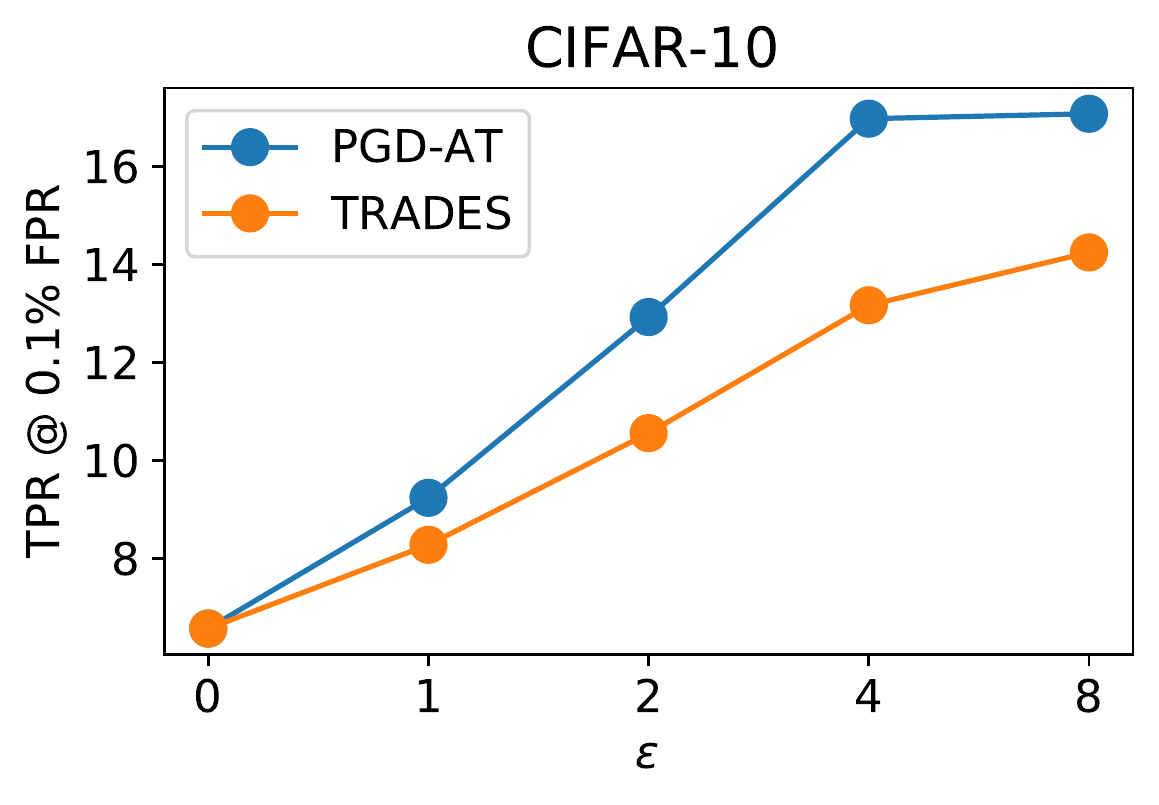}
  \caption{Attack success rates of PGD-AT and TRADES under different $\epsilon$ on CIFAR-10 with CNN-8.}
\label{fig.epstpr_cnn}
\end{figure}

The CNN-8 architecture used in this experiment is the same as the one employed for the SVHN dataset, as described in \cref{sec:setting} (Models paragraph). Furthermore, all hyper-parameter settings align with those described in \cref{sec:setting}. The evaluation results of different data augmentation methods are presented in \cref{fig.epstpr_cifar10_cnn}, which provides a concise representation of the results similar to those shown in \cref{tab:c10}. From this figure, we observe a less pronounced correlation between generalization gaps and privacy leakages compared to the previous MaxPreCA method. Regarding the adversarial training methods, the results are presented in \cref{tab:robustness_cifar10_cnn} and \cref{fig.epstpr_cnn}. Overall, these results closely resemble the trends observed with the ResNet-18 architecture on CIFAR-10 in \cref{tab:robustness} and \cref{fig.epstpr}. This consistency in trends supports that our findings remain consistent
across diverse model architectures and datasets.

\section{Membership Inference Attack Results on Non-image Datasets}
\label{supp:nonImg}
\renewcommand{\thefigure}{D.\arabic{figure}}
\setcounter{figure}{0}
\renewcommand{\thetable}{D.\arabic{table}}
\setcounter{table}{0}

\begin{table*}[!t]
  \centering
  \small
\renewcommand{\arraystretch}{1.15}{
  \setlength{\tabcolsep}{4pt}
  {
    \begin{tabular}{c|cc|cccc}
    \textbf{Method} & \textbf{Training Acc}  & \textbf{Test Acc} & \textbf{TPR @ 0.1\% FPR} & \textbf{TPR @ 0.001\% FPR} & \textbf{Log-scale AUC} & \textbf{MIA Balanced Acc} \\
    
     \hline
None & $100.0 \pm 0.0$ & $75.2 \pm 0.2$     & $19.34 \pm 1.09$ & $4.97 \pm 1.49$     & $0.881 \pm 0.005$     & $85.60 \pm 0.20$ \\
\hline
Disturblabel & $100.0 \pm 0.0$ & $75.7 \pm 0.3$     & $\bm{20.69} \pm 0.86$ & $4.55 \pm 3.10$     & $0.873 \pm 0.030$     & $84.60 \pm 0.14$ \\
Noise & $100.0 \pm 0.0$ & $75.3 \pm 0.3$     & $19.14 \pm 1.07$ & $3.85 \pm 1.71$     & $0.876 \pm 0.007$     & $85.58 \pm 0.19$ \\
Distillation & $100.0 \pm 0.0$ & $75.5 \pm 0.3$     & $\bm{21.54} \pm 1.54$ & $\bm{5.08} \pm 2.27$     & $\bm{0.885} \pm 0.009$     & $84.67 \pm 0.36$ \\
01 Flipping & $100.0 \pm 0.0$ & $75.1 \pm 0.4$     & $\bm{22.16} \pm 1.39$ & $4.90 \pm 1.90$     & $\bm{0.886} \pm 0.010$     & $\bm{86.64} \pm 0.12$ \\
    \end{tabular}
    }
    }
  \caption{Attack success rates of different data augmentation methods on Purchases. }
  \label{tab:purchase}
\end{table*}

\begin{table*}[!t]
  \centering
  \small
\renewcommand{\arraystretch}{1.15}{
  \setlength{\tabcolsep}{4pt}
  {
    \begin{tabular}{c|cc|cccc}
    \textbf{Method} & \textbf{Training Acc}  & \textbf{Test Acc} & \textbf{TPR @ 0.1\% FPR} & \textbf{TPR @ 0.001\% FPR} & \textbf{Log-scale AUC} & \textbf{MIA Balanced Acc} \\
    
     \hline
None & $100.0 \pm 0.0$ & $61.6 \pm 0.8$     & $52.43 \pm 9.57$ & $43.56 \pm 14.33$     & $0.960 \pm 0.014$     & $94.89 \pm 0.34$ \\
\hline
Disturblabel & $100.0 \pm 0.0$ & $63.4 \pm 0.8$     & $47.78 \pm 5.75$ & $36.42 \pm 9.27$     & $0.953 \pm 0.009$     & $91.95 \pm 0.42$ \\
Noise & $97.1 \pm 0.4$ & $64.1 \pm 0.7$     & $43.04 \pm 6.14$ & $33.36 \pm 7.15$     & $0.948 \pm 0.008$     & $92.08 \pm 0.37$ \\
Distillation & $100.0 \pm 0.0$ & $61.4 \pm 0.9$     & $\bm{53.89} \pm 8.19$ & $40.97 \pm 9.01$     & $\bm{0.961} \pm 0.008$     & $94.85 \pm 0.53$ \\
01 Flipping & $100.0 \pm 0.0$ & $65.7 \pm 0.8$     & $46.78 \pm 5.56$ & $30.42 \pm 12.39$     & $0.945 \pm 0.014$     & $92.12 \pm 0.54$ \\
    \end{tabular}
    }
    }
  \caption{Attack success rates of different data augmentation methods on Locations.}
  \label{tab:locations}
\end{table*}

We performed similar experiments with an MLP architecture on two additional non-image datasets, namely Purchases and Locations processed by \citet{shokri2017membership}\footnote{\url{https://github.com/privacytrustlab/datasets}}. The Purchases dataset is based on the ``acquire valued shoppers'' challenge dataset on Kaggle, containing shopping histories for several thousand individuals. \citet{shokri2017membership} derived a simplified purchase dataset with 197,324 data samples, where each sample consists of 600 binary features. The first 60,000 data samples from Purchases are
used to perform 100-category classification in our experiments. The Locations dataset is processed from the publicly available set of mobile users' location ``check-ins'', which contained 5,010 data samples with 446 binary features. All 5,010 data samples from Purchases are used to perform classification on 30-category classification in our experiments.

\renewcommand{\thefigure}{E.\arabic{figure}}
\setcounter{figure}{0}
\renewcommand{\thetable}{E.\arabic{table}}
\setcounter{table}{0}

\begin{figure}[!t]
  \centering
  \includegraphics[width=0.8\linewidth]{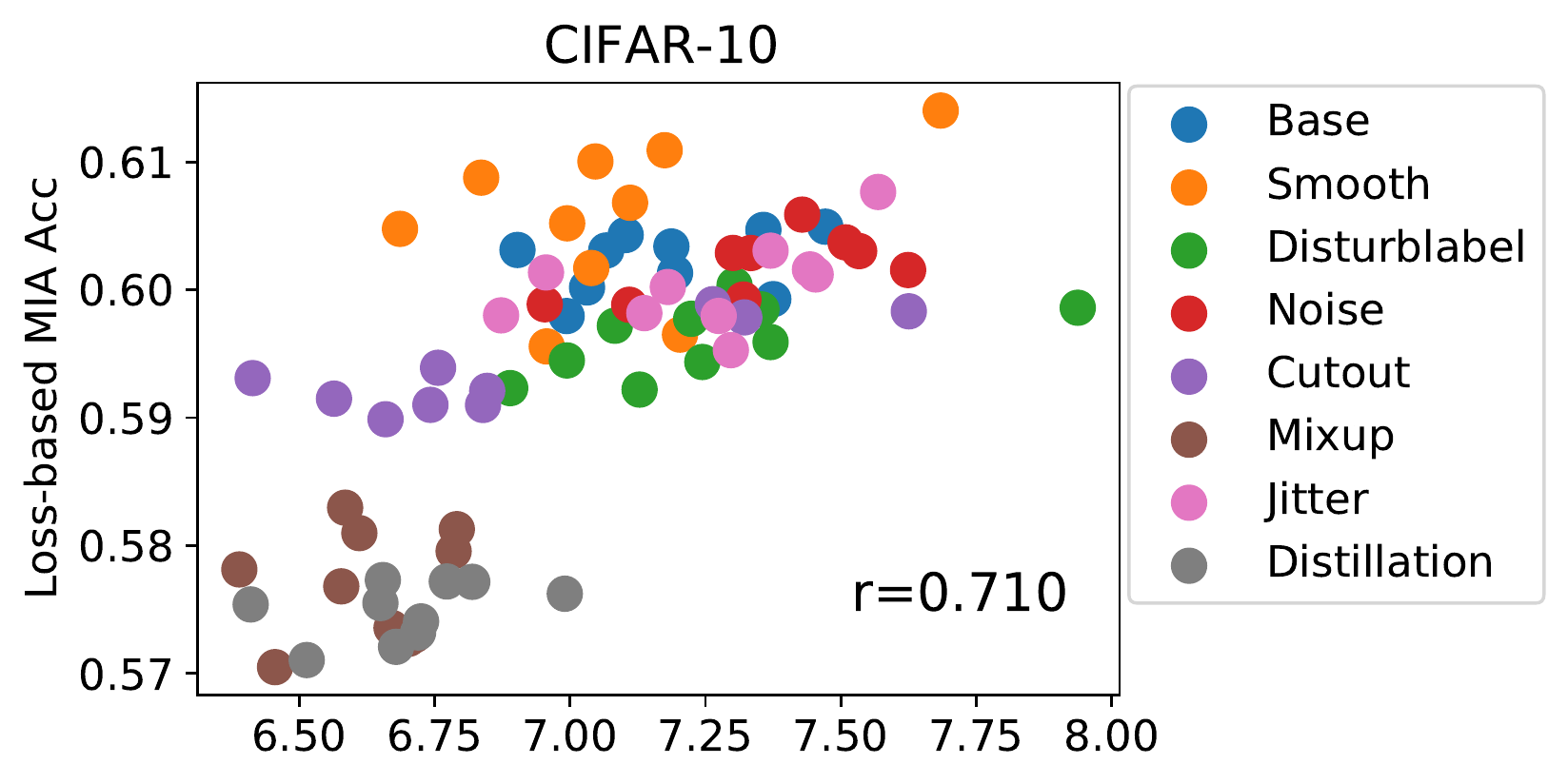}
  \caption{Attack success rate versus the train-test gap of different data augmentation models on CIFAR-10 using loss-based MIA \cite{yeom2018privacy}. $r$ stands for the Pearson correlation coefficient.}
\label{fig.epstpr_loss}
\end{figure}

\begin{figure}[!t]
  \centering
  \includegraphics[width=0.8\linewidth]{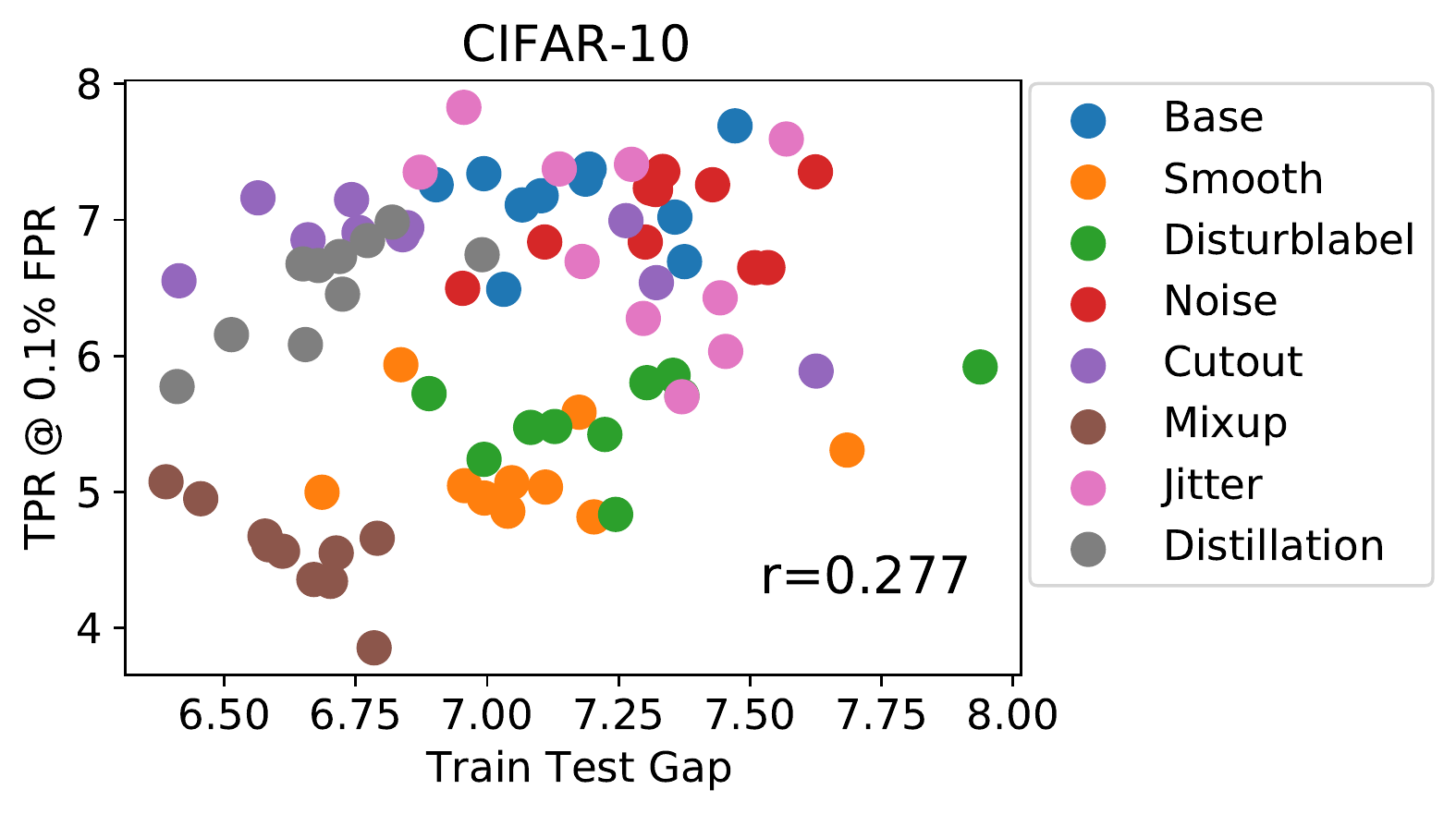}
  \caption{Attack success rate versus the train-test gap of different data augmentation methods on CIFAR-10 using difficulty calibrated loss approach \cite{importance}.}
\label{fig.epstpr_tausingle}
\end{figure}

The MLP architecture we used in the experiments has five fully connected layers, with sizes 512, 256, 128, 128, and 128 in turn. Batch normalization was applied to each layer. As this architecture is relatively simple, we only trained it for 50 epochs. Multi-step decay which scales the learning rate by 0.1 was used on the 25th and 37th epochs. Other experimental settings default to those described in \cref{sec:setting}.

As mentioned in \cref{sec:disscussions}, we investigated the privacy impacts of applying Disturblabel, Distillation, and Gaussian Noise. Additionally, as the features of both Purchases and Locations are binary, we investigated a new 0-1 Flipping augmentation method specially designed for these non-image datasets. Concretely, the 0-1 Flipping augmentation involves flipping a certain proportion of features, with 1\% of features flipped for Purchases and 10\% for Locations. The overall evaluation results on these datasets are shown in Tables \ref{tab:purchase} and \ref{tab:locations}.

\section{Membership Inference Attack Results Beyond LiRA}
\label{supp:tau}

\renewcommand{\thefigure}{E.\arabic{figure}}
\setcounter{figure}{0}
\renewcommand{\thetable}{E.\arabic{table}}
\setcounter{table}{0}

\begin{table}[!t]
  \centering
  \small
	\renewcommand{\arraystretch}{1.15}{
  \setlength{\tabcolsep}{4pt}
  {
    \begin{tabular}{c|c|cc}
	\multirow{2}*{\textbf{Dataset}} & \multirow{2}*{\textbf{Method}} & \textbf{Adversarial} & \textbf{TPR @}  \\
	 &  & \textbf{Acc} & \textbf{0.1\% FPR} \\
	\hline
	\multirow{5}*{\textbf{CIFAR-10}} 
        & Base & $0.0 \pm 0.0$ & $7.15 \pm 0.33$ \\
        \cline{2-4}
        & PGD-AT & $38.8 \pm 0.4$ & $15.42 \pm 1.20$ \\
	& TRADES & $45.2 \pm 0.3$ & $7.54 \pm 1.371$ \\
	& AWP & $45.9 \pm 0.1$ & $7.32 \pm 0.86$ \\
	& TRADES-AWP & $48.8 \pm 0.2$ & $7.54 \pm 0.73$ \\
	
    \end{tabular}
    
    }
   }
   \caption{The accuracies on adversarial examples (Adversarial Acc) and privacy leakage of different adversarial training models on CIFAR-10 using difficulty calibrated loss approach \cite{importance}. The accuracies are evaluated using PGD with $\epsilon = 8$ and 20 iteration steps.}
  \label{tab:robustness_newattack}
\end{table}

To demonstrate that the results using other MIAs discussed in \cref{subsec.ave} are similar to those using the MaxPreCA, \cref{fig.epstpr_loss} showcases the attack success rate versus the train-test gap of different data augmentation models on CIFAR-10 using loss-based MIA \cite{yeom2018privacy}. 

To demonstrate that our findings are not limited to LiRA alone, \cref{fig.epstpr_tausingle} and \cref{tab:robustness_newattack} showcase the results using the difficulty calibrated loss approach \cite{importance}. As discussed in \cref{sec:beyond}, these results indicate that the general trends in privacy effects resulting from data augmentation and adversarial training align with those observed using LiRA (see \cref{fig.epstpr} and \cref{tab:robustness}).

\end{document}